\documentclass[lettersize,journal]{IEEEtran}
\usepackage{algorithmic}
\usepackage{algorithm}
\usepackage{array}
\usepackage{textcomp}
\usepackage{stfloats}
\usepackage{url}
\usepackage{verbatim}
\usepackage{graphicx}
\usepackage{cite}
\usepackage{amsmath,amssymb,amsfonts}
\usepackage{bm} 
\usepackage{multirow} 
\usepackage{supertabular}
\usepackage[T1]{fontenc}
\usepackage{longtable} 
\usepackage{xcolor}
\usepackage{tabularx}
\usepackage{comment}
\usepackage{mathrsfs} 
\usepackage{float} 
\usepackage{booktabs}

\usepackage{hyperref}
\usepackage{subcaption}
\begin{document}

\title{Traffic-Context-Aware Driver Emotion Recognition with Expectation-Deviation-Driven Inference}

\author{Longhao Yan, Jinhao Liang, Kaidi Yang
\thanks{The authors are with the Department of Civil and Environmental Engineering, National University of Singapore, Singapore 117576. Email:{longhao.yan@u.nus.edu, jh.liang0523@gmail.com, kaidi.yang@nus.edu.sg}. \textit{Corresponding author}: Kaidi Yang. }
\thanks{This research was supported by the Singapore Ministry of Education (MOE) under its AcRF Tier 1 Grant (A-8000404-01-00). 
}}

\maketitle

\begin{abstract}
Driver emotion provides an important indication of how drivers perceive and evaluate ongoing traffic situations, making reliable emotion recognition an important capability for effective human-vehicle interaction. Yet, most existing driver emotion recognition approaches rely primarily on in-cabin images and overlook how surrounding traffic interactions shape driver emotional responses. To address this limitation, we propose a traffic-context-aware driver emotion recognition (TADER) framework that integrates in-cabin signals with out-of-cabin traffic scenarios, modeling driver emotions as arising from \emph{expectation deviation} represented by the discrepancy between predicted and observed vehicle trajectories.
The proposed framework comprises two key components: (1) a vision-based base recognizer that extracts spatiotemporal representations from in-cabin facial images to infer driver emotions, and (2) a traffic-context-aware fusion module that quantifies expectation deviation from external traffic scenarios and integrates it into the base recognizer via attention-based embedding fusion.
To overcome the scarcity of paired in-cabin and out-of-cabin data, we introduce a transfer learning strategy that transfers a trajectory prediction model trained on large-scale naturalistic datasets to smaller paired datasets.
Extensive experiments demonstrate that TADER consistently outperforms state-of-the-art methods across both naturalistic and simulated driving environments by incorporating traffic context information into emotion recognition, resulting in improvements of 5\% in accuracy on the AIDE and manD datasets. 
\end{abstract}

\begin{IEEEkeywords}
Driver Emotion Recognition, Multi-Modal Fusion, Human-Centered Automated Driving, Expectation-Deviation-Driven Inference
\end{IEEEkeywords}

\section{Introduction}

\IEEEPARstart{U}{nderstanding} driver states is essential for effective human-vehicle interaction in automated driving systems in which human drivers or safety operators remain involved in vehicle control, supervision, or takeover~\cite{zhu2024frequency,jiang2023efficient,an2022noncontact,tan2021human,carfi2021gesture}. 
Among various driver states, emotion provides important information about how individuals perceive and evaluate the driving situation and the automated system, as well as their readiness to respond. Emotion recognition is especially critical in automated driving systems under shared control, where operational authority and responsibility are dynamically distributed between the driver and the vehicle. Recognizing driver emotion can help automated driving systems infer the driver’s psychological state, perception of surrounding risks, and potential response intentions. Such information enables the system to anticipate driver needs and adapt its assistance strategy accordingly, thereby supporting safer and more human-centered interaction.

Most existing approaches formulate driver emotion recognition as a pattern recognition problem based on in-cabin signals such as facial expression, body posture, hand gestures, and speech \cite{yang2023robust,xia2020cross,du2020convolution,li2021cogemonet,li2021spontaneous,li2025glf,hsu2025cross,tao2024hierarchical,chen2024static,liu2024mmatrans,mao2024poster++,malik2025action,chen2022emotion}. 
Early research focuses almost exclusively on facial information. 
For example, Ref~\cite{yang2023robust} proposes a high-purity feature separation framework for robust driver emotion recognition in scenarios with heterogeneous drivers and illumination conditions. 
Ref~\cite{xia2020cross} proposes a cross-dataset driver emotion recognition method that exploits global discriminative information and local structural knowledge to address the distribution mismatch between offline training data and real-time on-road driving scenarios. 
However, these methods rely solely on facial expression analysis, which is highly sensitive to occlusion, lighting variation, and individual differences, thereby leading to frequent misclassifications.
Subsequent studies extend facial analysis by incorporating additional in-cabin signals.
Ref~\cite{du2020convolution} utilizes a Convolution Bidirectional Long Short-term Memory Neural Network to predict the driver’s emotion based on geometric features extracted from facial skin information and heart rate changes in RGB components. Ref~\cite{li2021cogemonet} proposes a biographical-feature-augmented driver emotion detection method to recognize driver emotion based on facial expression and personal characteristics, including age, gender, and driving experience.
Despite these advances, existing approaches tend to model correlations between emotion and in-cabin signals, while overlooking the complementary traffic context evidence contained in dynamic interactions among vehicles.

In driving scenarios, driver emotion is fundamentally shaped by the external traffic environment, as emotions emerge from continuous interactions with surrounding vehicles. To the best of our knowledge, only Ref. \cite{mou2023driver} incorporates external traffic information beyond the vehicle cabin, which develops a multimodal fusion framework based on static features from the driver’s eyes, the vehicle, and the surrounding environment. 
However, this work considers only the distance to the preceding vehicle and does not explicitly model the dynamic and time-evolving interactions between the ego vehicle and surrounding traffic, thereby limiting its ability to capture the traffic context in which driver emotions occur.

Existing cognitive studies have shown that emotional responses are closely associated with expectations about how situations will evolve~\cite{heffner2021emotion,ding2024expectation,heffner2025separable}. 
Specifically, emotional responses can be triggered when observed outcomes deviate from these expectations. In driving scenarios, such expectation deviations can be represented by discrepancies between predicted and observed vehicle trajectories. However, existing driver emotion recognition methods rarely exploit such insights.

\vspace{0.15em} \noindent \emph{\textit{Statement of Contribution}}. To address these research gaps, this paper introduces a traffic-context-aware driver emotion recognition framework, termed Traffic-Context-Aware Driver Emotion Recognition (TADER). The proposed framework integrates in-cabin driver signals with out-of-cabin observations of surrounding traffic and characterizes traffic context through expectation deviations between predicted and observed vehicle trajectories.
Our contribution is threefold. First, TADER is among the first attempts to explicitly incorporate dynamic traffic context information into driver emotion recognition, with a particular focus on expectation deviations in multi-vehicle interactions. Second, we characterize such expectation deviations using a vehicle trajectory predictor and integrate the resulting traffic context representation with in-cabin visual information through an attention-based embedding fusion mechanism for driver emotion recognition. Third, to address the scarcity of datasets that jointly align in-cabin visual data with external vehicle trajectories, we leverage transfer learning from trajectory prediction models trained on large-scale public driving datasets to improve generalization under limited paired data.

The remainder of this paper is organized as follows. 
Section~\ref{sec:Problem_Formulation} describes the problem formulation. 
Section~\ref{sec:Method} describes the details of our TADER method. 
Section~\ref{sec:Result} provides extensive experimental comparison and evaluation. 
Section~\ref{sec:Conclusion} concludes the paper and presents future work.

\section{Problem Formulation}
\label{sec:Problem_Formulation}

We consider a traffic scene involving an ego vehicle driven by a human driver, denoted by $\text{ego}$, and a set of observable surrounding vehicles, denoted by $\mathcal{M}$. Let $\mathcal{V}=\mathcal{M}\cup\{\text{ego}\}$ denote the set of all vehicles considered. 
Traffic scene information is characterized by the trajectories of the vehicles in $\mathcal{V}$, represented by $\{p_j\}_{j\in\mathcal{V}}$, where $p_j$ is the time series of longitudinal and lateral positions of vehicle $j$. 
Let graph $\mathcal{G}=(\mathcal{V},\mathcal{E})$ represent the interaction topology among these vehicles. An edge exists between two vehicles if they interact, for example, when they are successive in the same lane or adjacent in neighboring lanes. Here, we adopt a realistic assumption that the set of surrounding vehicles $\mathcal{M}$ evolves slowly over time and therefore omit its explicit time index. 

We discretize the considered time horizon into time intervals of length $\Delta T$ (e.g., 3 seconds), indexed by $k$, over which emotion recognition is performed. This is because driver emotion typically evolves over short temporal windows rather than at individual time steps. Each interval is assigned a discrete ground-truth emotion label $e_k$ from the set $\mathcal{E}$. For each interval $k$, we consider historical trajectories over a time horizon $\mathcal{H}_k$ and realized trajectories over a prediction horizon $\mathcal{P}_k$, where $\mathcal{H}_k$ and $\mathcal{P}_k$ are typically chosen to be successive in time. 
Let $ \tau_k=\{p_{jt}\}_{j\in\mathcal{V},\,t\in\mathcal{H}_k} $ denote the historical trajectories of all vehicles in $\mathcal{V}$, where $p_{jt}$ represents the two-dimensional position of vehicle $j$ at time $t$. For each vehicle $j\in\mathcal{V}$, let $ \sigma_{j,k}=[p_{jt}]_{t\in\mathcal{P}_k} $ denote its ground-truth trajectory over the prediction horizon $\mathcal{P}_k$, and let $ \sigma_k=[\sigma_{j,k}]_{j\in\mathcal{V}} $ denote the ground-truth trajectories of all vehicles, ordered consistently with the vehicle nodes in $\mathcal{G}$.

Driver state information, e.g., facial expressions and body posture, is extracted from in-cabin video streams $X_{\text{in-cabin}}$ captured by a driver-facing camera. The ego vehicle's position is measured by onboard localization devices such as GPS, and the positions of surrounding vehicles can be extracted from ego-vehicle onboard sensors (e.g., camera videos or LiDAR point clouds) or directly communicated via connected vehicle technologies.

Given the synchronized in-cabin observations $X_{\text{in-cabin},k}$, historical vehicle trajectories $\tau_k$, and the ground-truth trajectories $\sigma_k$ during the prediction horizon, the goal of this study is to recognize the driver's emotion $e_k\in\mathcal{E}$ by jointly exploiting observable in-cabin responses and the traffic interaction context in which those responses occur. For brevity, we consider a single time interval by default in the remainder of this chapter and omit the index $k$ when no ambiguity arises.

\section{Methodology} \label{sec:Method} 
We propose a traffic-context-aware driver emotion recognition (TADER) framework that explicitly incorporates out-of-cabin multi-vehicle interaction information into emotion recognition.
In this framework, \emph{expectation deviation}, defined as the mismatch between predicted and observed vehicle trajectories, is used to characterize traffic context changes associated with driver emotion.
Here, we use predicted trajectories to represent driver expectations for two reasons. 
First, driver emotion and behavior are inherently shaped by the motion patterns of surrounding vehicles, which are naturally captured by their trajectories. 
Second, drivers continuously form internal predictive models of how surrounding vehicles are likely to evolve, interpret ongoing traffic situations accordingly, and guide their actions. 
When the actual trajectories of vehicles deviate from these internal predictions, the resulting expectation deviation can provide contextual evidence of cognitive evaluations related to uncertainty, risk, or unexpected events that are reflected in emotional responses.

\begin{figure*}[htbp]
    \centering
	    \includegraphics[width=0.92\linewidth]{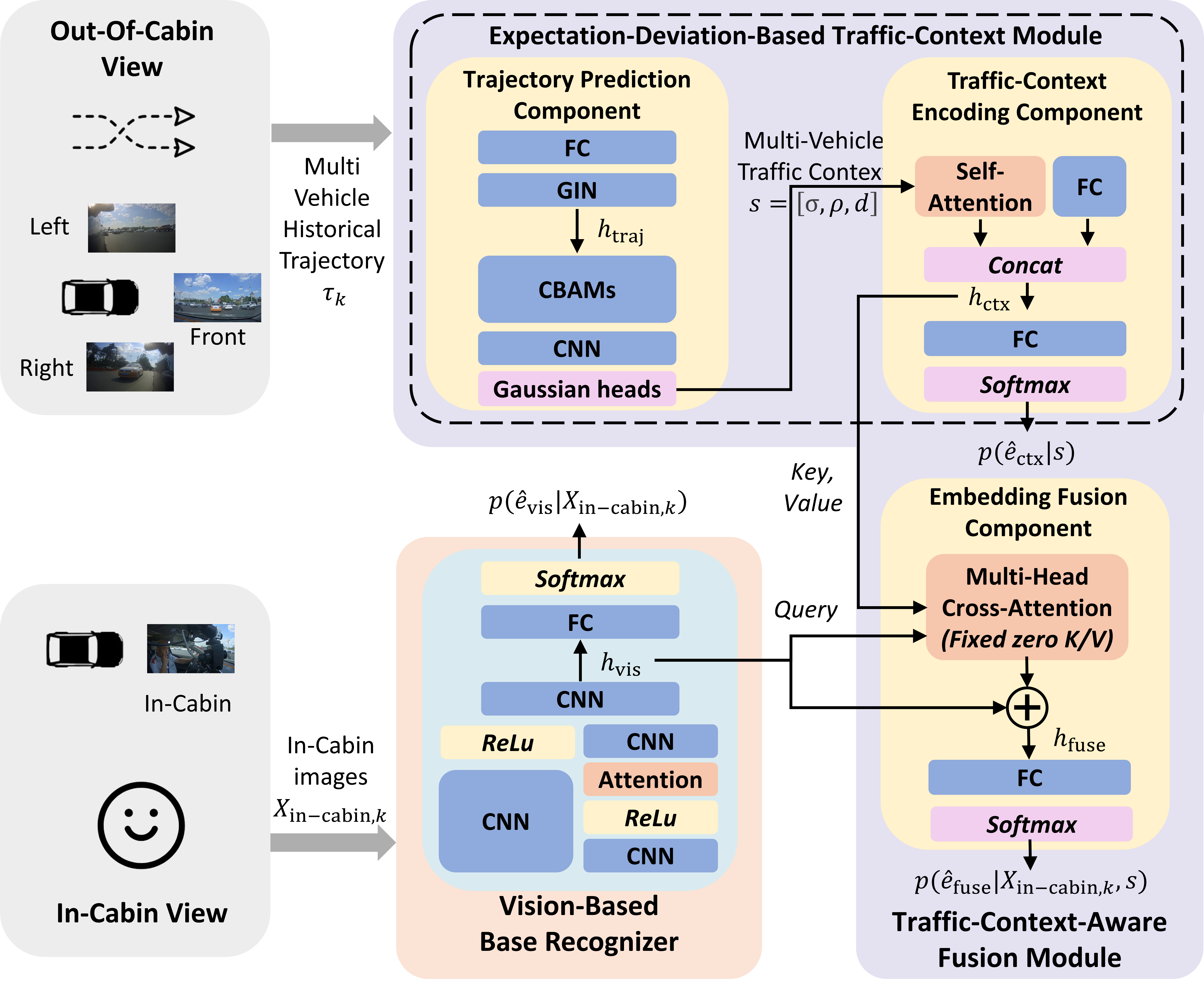}
    \caption{Framework of the traffic-context-aware driver emotion recognition method}
    \label{fig:CI_DER} 
\end{figure*}

As illustrated in Fig.~\ref{fig:CI_DER}, the proposed traffic-context-aware driver emotion recognition framework consists of two main components: (1) a vision-based base recognizer that extracts spatiotemporal representations from in-cabin driver facial images to provide a visual signal over emotional states (see Section~\ref{ssec:vision_based}) and (2) a traffic-context-aware module that quantifies the expectation deviation from out-of-cabin traffic scenes and integrates such deviation information into the base recognizer via attention-based embedding fusion (see Section~\ref{ssec:context_based}). 
Moreover, we address a key practical challenge in implementing the proposed framework, i.e., the limited availability of datasets that jointly align in-cabin visual data with external vehicle trajectories. To this end, we develop a transfer learning-based strategy that first learns a generalizable trajectory prediction model from large-scale open naturalistic datasets and then transfers the learned representations to a small dataset containing paired in-cabin and out-of-cabin data, as detailed in Section~\ref{ssec:transfer}.

\subsection{Vision-Based Base Recognizer} \label{ssec:vision_based}
The base module of our framework is a vision-based emotion recognizer that directly infers driver emotion from in-cabin driver-facing camera data. 
This choice reflects the prevailing pattern recognition paradigm in emotion recognition 
relying solely on visual information.
Specifically, we adopt the Context-Aware Emotion Recognition Network (CAERNet)~\cite{lee2019context}, which jointly models facial expressions and background information from in-cabin images. Note that CAERNet is selected for its strong empirical performance, and the proposed methodological framework is readily applicable to other vision-based emotion recognition architectures. 

At each time interval $k$, CAERNet decomposes in-cabin driver-facing camera data $X_{\text{in-cabin},k}$ into a facial stream $X_{\text{in-cabin},k}^{\text{face}}$ and a background stream $X_{\text{in-cabin},k}^{\text{bg}}$. The facial stream is used to recognize facial expressions, and the background stream captures auxiliary information such as body posture and gesture. 

CAERNet adopts a dual-branch encoder based on three-dimensional convolutional neural networks (CNNs) to extract spatiotemporal representations from both facial and background streams, i.e.,
\begin{align}
    \tilde{F}^{c} = \textit{ReLU}(\text{CNN}_{\theta_c}\left(X_{\text{in-cabin},k}^{c})\right),~c\in\{\text{face},\text{bg}\}
\end{align}
An attention mechanism is then used to emphasize informative spatial regions in the background stream, i.e., 
\begin{equation}
    \hat{F}^{\text{bg}} = \text{CNN}_\psi\left(\sum_l \hat{A}_l \tilde{F}^{\text{bg}}_l\right),
\end{equation}
where $\hat{A}_l = \frac{\exp(A_l)}{\sum_m \exp(A_m)}$ denotes the normalized attention weight at pixel $l$, representing the relative importance of each spatial feature in $\tilde{F}^{\text{bg}}$.

The extracted facial features $\tilde{F}^{\text{face}}$ and refined background features $\hat{F}^{\text{bg}}$ are then fused in the decoder using a fusion mechanism with learnable weights. 
This mechanism calculates a weighted concatenation $C= \text{Concat}\left(\Gamma_{\text{face}} \tilde{F}^{\text{face}},\, \Gamma_{\text{bg}} \hat{F}^{\text{bg}}\right)$, where $\Gamma_{\text{face}}$ and $\Gamma_{\text{bg}}$ are learnable coefficients to balance the contributions of the facial and background modalities. 
This concatenation $C$ is then processed by CNN blocks to produce the final visual feature embedding $h_{\text{vis}}$:
\begin{equation}
    h_{\text{vis}}=\text{CNN}_{\text{dec}}(C)
\end{equation}

The emotion distribution is then predicted as
\begin{equation}
    p(\hat{e}_{\text{vis}}|X_{\text{in-cabin},k}) = \textit{Softmax}(W_v h_{\text{vis}} + b_v),
\end{equation}
where $W_v$ and $b_v$ denote learnable parameters.

Finally, we define a vision-based cross-entropy (CE) loss with respect to the ground-truth emotion label $e$:
\begin{equation}
    \mathcal{L}_{\text{vis}} = \mathrm{CE}(p(\hat{e}_{\text{vis}}|X_{\text{in-cabin},k}), e),
\end{equation}

\subsection{Traffic-Context-Aware Fusion Module} \label{ssec:context_based}
In this subsection, we introduce a traffic-context-aware fusion module that augments the base recognizer, which corresponds to prevailing vision-based emotion recognizers relying solely on driver-facing camera data. 
Specifically, this module characterizes traffic context through discrepancies between predicted and observed multi-vehicle dynamics.
This module comprises two components: 
(i) an expectation-deviation-based traffic context feature extraction component that represents traffic context using discrepancies between predicted and observed vehicle trajectories (Section~\ref{sssec:exp_modeling}) and
(ii) an attention-based embedding fusion component that integrates traffic context features with visual embeddings from driver-facing cameras to produce coherent emotion recognition outputs (Section~\ref{sssec:fusion}).

\subsubsection{Expectation-Deviation-Based traffic context Feature Extraction} \label{sssec:exp_modeling}
This module extracts traffic context features through two key components: (i) expectation deviation, which is quantified by comparing predicted and observed vehicle trajectories, and (ii) traffic context encoding, which maps expectation deviations to embeddings and, in turn, to recognized driver emotions.

\noindent \textbf{(i) Expectation deviation quantification}. \label{para:exp_model} 
This component quantifies expectations by predicting vehicle trajectories based on the out-of-cabin traffic context, derived by modeling spatiotemporal interactions among vehicles.
Accordingly, we design a trajectory prediction model based on the Pishgu architecture~\cite{alinezhad2023pishgu}, which captures interaction-aware motion dynamics among vehicles to generate expected trajectories. 
We extend this model to predict the future trajectory of every vehicle $j\in\mathcal{V}$ by producing a probabilistic trajectory distribution from the joint historical observations.
Notice that the predictor can be instantiated by any architecture that models the surrounding dynamics as interaction graphs to predict future trajectories.

Using the interaction graph $\mathcal{G}$, historical trajectories $\tau_k$, and ground-truth trajectories $\sigma_k$ defined in Section~\ref{sec:Problem_Formulation}, we characterize the expectation as the predicted trajectories over $\mathcal{P}_k$ conditioned on the historical trajectories over $\mathcal{H}_k$. To better capture motion dynamics, let $\tau_{\text{rel},k}$ denote the relative position increments of the vehicles between consecutive time steps.

Using the constructed graph representation, we employ a Graph Isomorphism Network (GIN) to capture interactions among agents by encoding their flattened historical trajectories. Specifically, the GIN aggregates messages over the graph $\mathcal{G}$ to produce 
the node-wise interaction-aware representation $H_{\text{traj},k}$, whose elements correspond to the vehicle nodes in $\mathcal{V}$:
\begin{equation}
    H_{\text{traj},k}
    =
    \text{GIN}\Big(
    \text{FC}(\text{Concat}(\tau_k,\tau_{\text{rel},k})),
    \mathcal{G}
    \Big).
\end{equation}
Subsequently, we fuse $H_{\text{traj},k}$ with the historical trajectories $\tau_k$ via channel-wise concatenation to preserve both interaction contexts and motion details:
$z_{0}=\text{Concat}\left(H_{\text{traj},k},\tau_k\right)$.

The combined feature $z_{0}$ is then refined through a sequence of convolutional blocks $\text{Conv2d}$. Each block incorporates a Convolutional Block Attention Module (CBAM) to sequentially apply channel and spatial attention, thereby emphasizing key motion-related patterns:
\begin{equation}
    z_{o} = \text{CBAM}\left( \text{Conv2d}(z_{o-1}) \right),
\end{equation}
where $o \in \{1, ..., O\}$ denotes the block index. This hierarchical structure enables a progressive extraction of important spatio-temporal dependencies from the fused representation.
The extracted representation $z_O$ is used to 
generate vehicle-specific Gaussian parameters for the future trajectories of all vehicles in $\mathcal{V}$:
\begin{align}
    \mu_{\hat{\sigma}_j}
    &=
    [\text{Conv2d}_{\mu}(z_O)]_j, \\
    \Sigma_{\hat{\sigma}_j}
    &=
    \operatorname{Diag}\!\left(
    \textit{softplus}\!\left(
    [\text{Conv2d}_{\Sigma}(z_O)]_j
    \right)
    +
    \varepsilon
    \right),
    \qquad j\in\mathcal{V}.
\end{align}
where $\varepsilon$ is a small positive constant added for numerical stability.
The resulting set of distributions
$\{\mathcal{N}(\mu_{\hat{\sigma}_j},\Sigma_{\hat{\sigma}_j})\}_{j\in\mathcal{V}}$
defines the predicted future trajectories of the ego vehicle and its surrounding vehicles.
For each vehicle $j\in\mathcal{V}$, its expectation deviation and prediction confidence are defined as
\begin{equation}
    d_j
    =
    \sigma_j-\mu_{\hat{\sigma}_j},
    \qquad
    \rho_j
    =
    \exp\left(
    -\tfrac{1}{2}
    d_j^\top
    \Sigma_{\hat{\sigma}_j}^{-1}
    d_j
    \right).
    \label{eq:likelihood}
\end{equation}
where $d_j$ is the expectation deviation associated with vehicle $j$, and $\rho_j$ denotes the prediction confidence quantified by the covariance-normalized agreement between the observed and predicted trajectories.
The multi-vehicle quantities are then organized according to the node order in $\mathcal{G}$:
\begin{equation}
    \sigma=[\sigma_j]_{j\in\mathcal{V}},
    \qquad
    \rho=[\rho_j]_{j\in\mathcal{V}},
    \qquad
    d=[d_j]_{j\in\mathcal{V}},
\end{equation}
and the traffic context representation is constructed as
\begin{equation}
    s=[\sigma,\rho,d],
\end{equation}
which jointly models the actual multi-vehicle trajectories, vehicle-specific prediction confidence, and multi-vehicle expectation deviations.

\noindent \textbf{(ii) Traffic context encoding}.
Building on the multi-vehicle deviation-based traffic context representation $s$, we introduce a traffic context encoding component that fuses the actual trajectories $\sigma$, vehicle-specific prediction confidence $\rho$, and multi-vehicle expectation deviations $d$ to produce a compact contextual representation. Specifically, self-attention is applied over the expectation-deviation sequences $d$ to capture key temporal patterns and dependencies across vehicles, while the actual trajectories $\sigma$ and prediction confidence $\rho$ are transformed via a fully connected layer. The resulting representations are concatenated to form the final traffic context embedding:
\begin{equation}
    h_{\text{ctx}}
    =
    \text{Concat}\big(
    \text{SelfAttn}(d),
    \text{FC}([\sigma,\rho])
    \big).
\end{equation}

We then compute the emotion distribution from the resulting multi-vehicle traffic context embedding:
\begin{equation}
    p(\hat{e}_{\text{ctx}}|s) = \textit{Softmax}(W_c h_{\text{ctx}} + b_c)
\end{equation}
where $W_c$ and $b_c$ are learnable parameters.

The traffic-context-based CE loss is accordingly defined for the expectation-deviation-based module:
\begin{equation}
    \mathcal{L}_{\text{ctx}} = \mathrm{CE}(p(\hat{e}_{\text{ctx}}|s),\, e) \label{eq:loss_ctx}
\end{equation} 

\subsubsection{Attention-Based Embedding Fusion} \label{sssec:fusion} 
After extracting expectation-deviation-based 
traffic context embedding, we formulate emotion recognition as a traffic-context-aware representation learning problem that jointly exploits traffic context information and in-cabin visual information.

This formulation is implemented by residual multi-head cross-attention, in which the visual embedding $h_{\text{vis}}$ serves as the query and the traffic context embedding $h_{\text{ctx}}$ provides the key and value.
To enable adaptive attention weighting between these two embeddings, a fixed zero key--value pair is appended after projection in each attention head, with both positions remaining unmasked. This allows the visual query to regulate the contribution of traffic context information.
Specifically, for each attention head $\ell$, the projected query, key, and value are given by $q_\ell=W_\ell^Qh_{\text{vis}}+b_\ell^Q$, $k_\ell=W_\ell^Kh_{\text{ctx}}+b_\ell^K$, and $v_\ell=W_\ell^Vh_{\text{ctx}}+b_\ell^V$, respectively.
The resulting fused embedding is computed as
\begin{equation}
\begin{aligned}
h_{\text{fuse}}
&= h_{\text{vis}} + W_O\text{Concat}
\left(
a_1v_1,\ldots,a_{N_h}v_{N_h}
\right)+b_O,
\end{aligned}
\label{eq:residual_fusion}
\end{equation}
where $ a_\ell = \frac{\exp(q_\ell^\top k_\ell/\sqrt{d_h})} {1+\exp(q_\ell^\top k_\ell/\sqrt{d_h})} $ denotes the attention weight assigned to the traffic context in head $\ell$, $N_h$ denotes the number of attention heads, $d_h$ is the feature dimension of each head, and $W_O$ and $b_O$ are learnable parameters that project the attention output to the dimension of $h_{\text{vis}}$.
The attention-weighted traffic context information is then integrated with the visual embedding through the residual connection, allowing both modalities to contribute to the final emotion prediction.

The fused embedding $h_{\text{fuse}}$ is then used to predict the emotion distribution:
\begin{equation}
p(\hat{e}_{\text{fuse}}\mid X_{\text{in-cabin},k},s) = \textit{Softmax}(W_fh_{\text{fuse}} + b_f),
\end{equation}
where $W_f$ and $b_f$ are learnable parameters.

A fusion-based CE loss is defined for the attention-based embedding fusion component to align the fused emotion prediction with the ground truth:
\begin{equation}
\mathcal{L}_{\text{fuse}} = \mathrm{CE}(p(\hat{e}_{\text{fuse}}|X_{\text{in-cabin},k},s), e) \label{eq:loss_fuse}
\end{equation}

Finally, the overall training loss for the emotion recognition task is defined as the combination of vision-based, traffic-context-based, and fusion-based losses. 
\begin{equation}
\mathcal{L}_{\text{emo}} =
    \lambda_{\text{vis}} \mathcal{L}_{\text{vis}} +
    \lambda_{\text{ctx}} \mathcal{L}_{\text{ctx}} +
    \lambda_{\text{fuse}} \mathcal{L}_{\text{fuse}} \label{eq:loss_emo}
\end{equation}
This formulation jointly supervises the visual, traffic context, and fused emotion predictions. The weights $\lambda_{\text{vis}}, \lambda_{\text{ctx}}, \lambda_{\text{fuse}}$ balance the contributions of the respective loss components.

\subsection{Transfer Learning for Limited Paired Data} \label{ssec:transfer} 
Training the traffic-context-aware model presents a fundamental challenge in data acquisition, as it requires paired observations of in-cabin driver states and out-of-cabin traffic scenes, which are costly to collect at scale.
In contrast, large-scale open naturalistic driving datasets are widely available and provide rich coverage of diverse traffic scenes. 
Moreover, drivers’ interaction patterns with surrounding traffic exhibit strong regularities across contexts, enabling behavioral generalization beyond specific drivers or datasets.
Therefore, we leverage transfer learning that extracts generalizable trajectory representations from large-scale naturalistic driving datasets and subsequently aligns them with limited paired data for traffic-context-aware emotion recognition. 
Note that the transfer learning strategy can also be applied to augment in-cabin driver-facing camera data, for which we omit for brevity. 

Specifically, we implement transfer learning mainly for trajectory prediction. To this end, we maintain two trajectory prediction models (see Section~\ref{para:exp_model}): one for the source domain (i.e., a large-scale naturalistic driving dataset, denoted as $\mathcal{S}$) capturing generalizable driving behaviors, and one for the target domain (i.e., a small dataset containing paired driver states and traffic scenes, denoted as $\mathcal{A}$). Both models are initialized with the trajectory predictor pretrained on the source domain $\mathcal{S}$. 
During the training of the emotion recognition framework, the source-domain model remains fixed, while the target-domain model is updated using the limited paired data in the target domain, with two complementary objectives: (1) fitting the predicted trajectories to the ground-truth trajectories in the target domain using a Gaussian negative log-likelihood (NLL) loss, and (2) aligning the predicted trajectory distributions of the source and target models by minimizing their Kullback-Leibler (KL) divergence. 
\begin{equation} \label{eq:loss_traj_underbrace}
\begin{split}
    \mathcal{L}_{\text{traj}}
    &=
    \underbrace{
    -\log\,p_{\mathcal{A}}(\sigma\mid\tau,\mathcal{G})
    }_{\text{Target-domain fit}}
    +
    \alpha\,
    \underbrace{
    D_{\mathrm{KL}}\!\left(
    p_{\mathcal{S}}(\hat{\sigma}\mid\tau,\mathcal{G})
    \,\|\,
    p_{\mathcal{A}}(\hat{\sigma}\mid\tau,\mathcal{G})
    \right)
    }_{\text{Domain alignment}}.
\end{split}
\end{equation}
where $\alpha$ controls the contribution of the domain alignment term.

To sum up, the overall training loss of all modules is:
\begin{equation}
\mathcal{L}_{\text{total}} =
    \lambda_{\text{traj}} \mathcal{L}_{\text{traj}} + \mathcal{L}_{\text{emo}} \label{eq:loss_total}
\end{equation}
where $\lambda_{\text{traj}}$ balances the contribution of the trajectory prediction loss.

Our proposed method is summarized in Algorithm~\ref{alg:emotion_recognition}.

\begin{algorithm}[t]
\caption{Traffic-Context-Aware Driver Emotion Recognition with Expectation Deviation}
\label{alg:emotion_recognition}
\begin{algorithmic}[1]

\REQUIRE Multi-modal dataset with in-cabin video streams $X_{\text{in-cabin}}$ and
synchronized traffic scene information; trajectory prediction models pretrained
on large-scale naturalistic driving datasets

\ENSURE Emotion Recognition $p(\hat{e}_{\text{fuse}}|X_{\text{in-cabin},k},s)$

\FOR{each mini-batch of data samples}

    \STATE Extract facial representation $\tilde{F}^{\text{face}}$ and background
    representation $\hat{F}^{\text{bg}}$ from in-cabin images

    \STATE Fuse representations
    $C=\text{Concat}\left(
    \Gamma_{\text{face}}\tilde{F}^{\text{face}},
    \Gamma_{\text{bg}}\hat{F}^{\text{bg}}
    \right)$
    and get visual embedding $h_{\text{vis}}$

    \STATE Compute emotion distribution using visual embedding
    $p(\hat{e}_{\text{vis}}|X_{\text{in-cabin},k})$

    \STATE Predict the vehicle-specific future-trajectory distributions
    $\{\mathcal{N}(\mu_{\hat{\sigma}_j},
    \Sigma_{\hat{\sigma}_j})\}_{j\in\mathcal{V}}$
    from the joint historical trajectories $\tau_k$
    and interaction graph $\mathcal{G}$

    \STATE For each $j\in\mathcal{V}$, compute
    $d_j=\sigma_j-\mu_{\hat{\sigma}_j}$ and
    $\rho_j=\exp(-\tfrac{1}{2}
    d_j^\top\Sigma_{\hat{\sigma}_j}^{-1}d_j)$,
    and organize them as
    $\sigma=[\sigma_j]_{j\in\mathcal{V}}$,
    $\rho=[\rho_j]_{j\in\mathcal{V}}$, and
    $d=[d_j]_{j\in\mathcal{V}}$

    \STATE Form traffic context representation
    $s=[\sigma,\rho,d]$ and encode it into
    the traffic context embedding $h_{\text{ctx}}$

    \STATE Compute emotion distribution using traffic context embedding
    $p(\hat{e}_{\text{ctx}}|s)$

                    \STATE Obtain $h_{\text{fuse}}$ using residual multi-head
    cross-attention with a fixed zero key--value pair in each
    head. 
    \STATE Compute emotion distribution using fused embedding
    $p(\hat{e}_{\text{fuse}}|X_{\text{in-cabin},k},s)$

    \STATE Compute the total loss $\mathcal{L}_{\text{total}}$
    and update trainable parameters via backpropagation

\ENDFOR
\end{algorithmic}
\end{algorithm}

\section{Results}
\label{sec:Result}

In this section, we conduct experiments to evaluate the proposed TADER framework on two multimodal driver emotion datasets.
Section~\ref{sec:benchmark} introduces the benchmark methods, Section~\ref{sec:implement} details the experimental setup and implementation details, Section~\ref{sec:testing} reports the emotion recognition performance and analyzes the contributions of the visual and traffic context branches, Section~\ref{sec:ablation} presents ablation studies, and Section~\ref{sec:sensitivity} performs sensitivity analyses. 

\subsection{Benchmarks}
\label{sec:benchmark}

The following emotion recognition benchmarks are compared to evaluate the proposed method:

\begin{itemize}
    \item MTCNN-D~\cite{zhang2016joint}. This benchmark combines the Multi-Task Cascaded Convolutional Networks (MTCNN) framework with an LSTM module to model temporal dynamics across sequential facial frames. It represents a conventional CNN--LSTM dynamic facial emotion recognition baseline without traffic context information.
    \item MAE~\cite{sun2023mae}. This benchmark implements a self-supervised dynamic facial expression recognition (FER) method based on masked autoencoders. A local-global interaction transformer is employed to enhance static appearance reconstruction with explicit temporal facial motion modeling. This benchmark represents FER methods that leverage large-scale pretraining to improve performance.
    \item IAL~\cite{li2023intensity}. This benchmark implements a transformer-based method for FER, which introduces the intensity-aware loss to distinguish the image samples with relatively low expression intensities. This benchmark represents a state-of-the-art FER method that relies solely on facial features for emotion recognition.
    \item CAERNet~\cite{lee2019context}. This benchmark implements a context-aware FER approach that jointly leverages facial expressions and background information to enhance emotion recognition. It employs an attention mechanism that masks facial regions to emphasize background information. This method represents FER approaches that consider the background information of images (e.g., driver posture) rather than focusing solely on facial features.
    \item TADER (Ours). This benchmark implements our proposed traffic-context-aware emotion recognition framework. 
        The visual branch is instantiated using MTCNN-D, MAE, IAL, or CAERNet in the benchmark comparisons, while CAERNet serves as the default backbone for the subsequent ablation and sensitivity analyses.
    By jointly modeling in-cabin visual information and multi-vehicle traffic context information, TADER captures emotion through an explicit modeling of traffic context, enabling a comprehensive and situationally grounded understanding of driver emotion in complex driving environments.
\end{itemize}

The comparison between TADER and the visual-only benchmarks evaluates whether traffic context provides additional information beyond facial expression modeling. In particular, the matched comparison between CAERNet and TADER separates the contribution of the traffic context branch from differences in visual-backbone capacity.

\subsection{Experiment Setup and Implementation Details}
\label{sec:implement}

\begin{figure}[htbp]
    \centering

    \begin{subfigure}{\linewidth}
        \centering
        \includegraphics[width=0.96\linewidth]{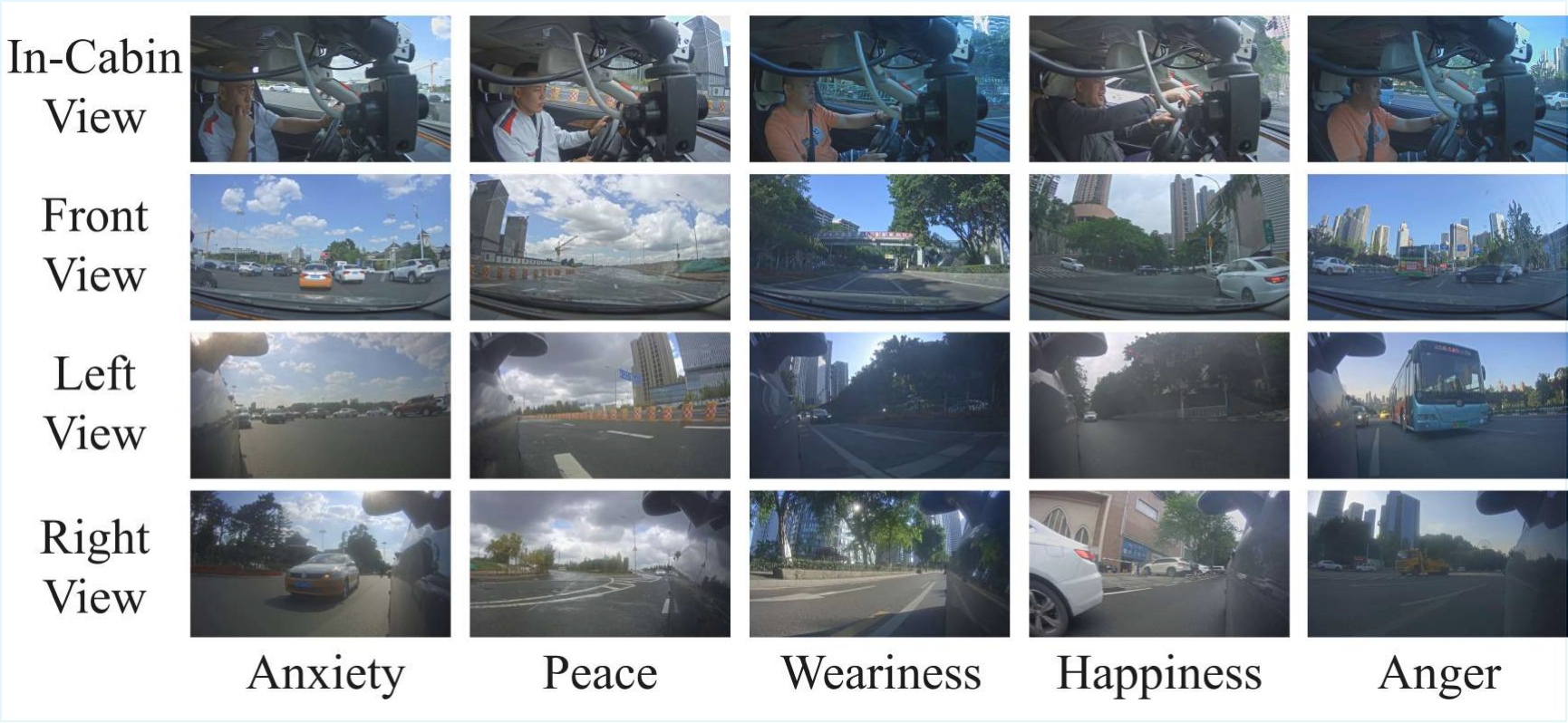}
        \caption{Representative samples of the AIDE dataset.}
        \label{fig:AIDE_Des_sub}
    \end{subfigure}

    \vspace{0.6em}

    \begin{subfigure}{\linewidth}
        \centering
        \includegraphics[width=0.96\linewidth]{2_manD_describe.pdf}
        \caption{Representative samples of the manD dataset.}
        \label{fig:manD_Des_sub}
    \end{subfigure}

    \caption{Illustration of the two multimodal driver emotion datasets used in this work.
    (a) AIDE captures naturalistic on-road driving with synchronized multi-view videos and five emotion categories.
    (b) manD contains controlled driving-simulator observations with synchronized facial video and vehicle and environment states labeled with five emotional states.}
    \label{fig:data_sources}
\end{figure}

\subsubsection{Datasets}

We utilize two multimodal driver emotion datasets: the AssIstive Driving pErception (AIDE) dataset~\cite{yang2023aide} and manD 1.0~\cite{dargahi2024multimodal} to evaluate our framework across both naturalistic on-road driving scenarios and simulated laboratory driving conditions, thereby demonstrating its robustness and applicability. 
\begin{itemize}
\item \textbf{AIDE} captures naturalistic on-road driving conditions with synchronized multi-view camera recordings (in-cabin, front, left, and right), providing paired observations of driver states and traffic scenes. Each AIDE data sample consists of a 3-second driving clip (45 frames per view), annotated with an emotion category $e \in \mathcal{E}=\{\textit{Anxiety},~\textit{Peace},\textit{Weariness}, \textit{Happiness}, \textit{Anger}\}$. Vehicle trajectories are obtained via YOLO-based detection and multi-object tracking~\cite{Jocher_Ultralytics_YOLO_2023}, projected into a bird’s-eye-view (BEV) coordinate system. Only lane-changing and forward-driving scenes are included, as they are both behaviorally expressive and emotionally informative.
\item \textbf{manD} is collected in a laboratory-based driving simulator, allowing controlled elicitation and repeatable measurement of emotional responses. It provides synchronized driver facial video, vehicle status signals, and environment perception measurements. We retain the emotion-induction phases of the emotional scenarios and the no-emotion scenario, assigning the corresponding condition labels $e \in \mathcal{E} = \{\textit{Fear}, \textit{Sadness}, \textit{Surprise}, \textit{Anger}, \textit{No Emotion}\}$ to the resulting clips.
\end{itemize}

To ensure consistent evaluation across datasets, we process AIDE and manD into a unified representation: each sample is segmented into a 3-second clip at 15 Hz, facial features are extracted from the in-cabin or driver-facing view, and vehicle motions are transformed into a BEV trajectory sequence aligned frame-by-frame with the visual inputs. For trajectory representation, we retain 10 observed steps and 35 future steps to characterize the driving context and behavioral expectations. 
Both datasets are split into 70\% training, 15\% validation and 15\% testing. To mitigate class imbalance and ensure consistent learning conditions across datasets, we perform class balancing separately on AIDE and manD to obtain an approximately uniform distribution over emotion categories. 
Fig.~\ref{fig:data_sources} presents representative examples from both datasets.

\subsubsection{Implementation Details}

We implement all benchmark methods using the default hyperparameter settings reported in their original publications, with minor adaptations to align their outputs with the five emotion categories defined in AIDE and manD.

For our proposed method, the visual branch processes 16 frames for each interval. The trajectory branch uses $T_o=10$ historical steps and $T_p=35$ future steps, with at most 12 vehicles represented in the interaction graph. The traffic context encoders contain 128 hidden units, while the internal trajectory predictor contains 64 hidden units. The trajectory predictor is pretrained on the large-scale naturalistic driving dataset NGSIM~\cite{alexiadis2004next} and subsequently adapted to AIDE and manD using the transfer learning strategy described in Section~\ref{ssec:transfer}.
The highD dataset~\cite{krajewski2018highd} is used as an alternative source dataset in the source-domain sensitivity analysis.

All models are optimized for 50 epochs using Adam with a learning rate of $10^{-4}$, a weight decay of $10^{-4}$, and a batch size of 40. The trajectory-loss weight $\lambda_{\text{traj}}$ is set to 0.05, and the source-target alignment coefficient $\alpha$ is set to 0.5. All models are implemented in PyTorch and trained on a high-performance computing platform equipped with NVIDIA H100 and A40 GPUs.

\subsection{Performance Analysis}
\label{sec:testing}

\begin{figure*}[htbp]
    \centering

        \begin{subfigure}[t]{0.31\textwidth}
        \centering
        \includegraphics[width=\linewidth]{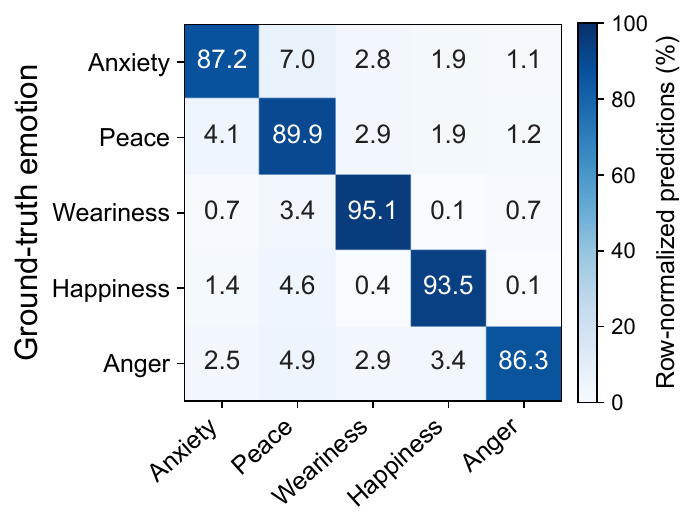}
        \caption{Visual branch on AIDE}
        \label{fig:confuse_face_aide}
    \end{subfigure}
    \hfill
    \begin{subfigure}[t]{0.31\textwidth}
        \centering
        \includegraphics[width=\linewidth]{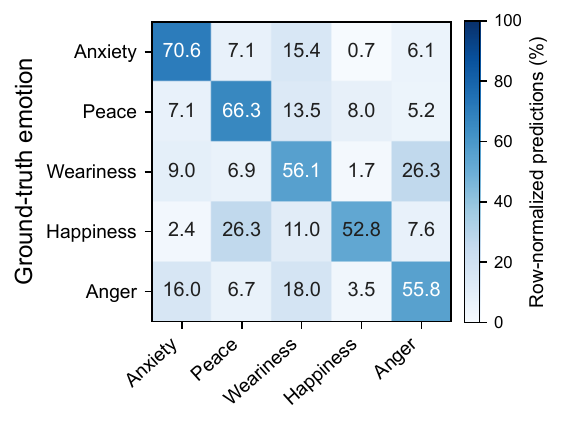}
        \caption{Traffic context branch on AIDE}
        \label{fig:confuse_context_aide}
    \end{subfigure}
    \hfill
    \begin{subfigure}[t]{0.31\textwidth}
        \centering
        \includegraphics[width=\linewidth]{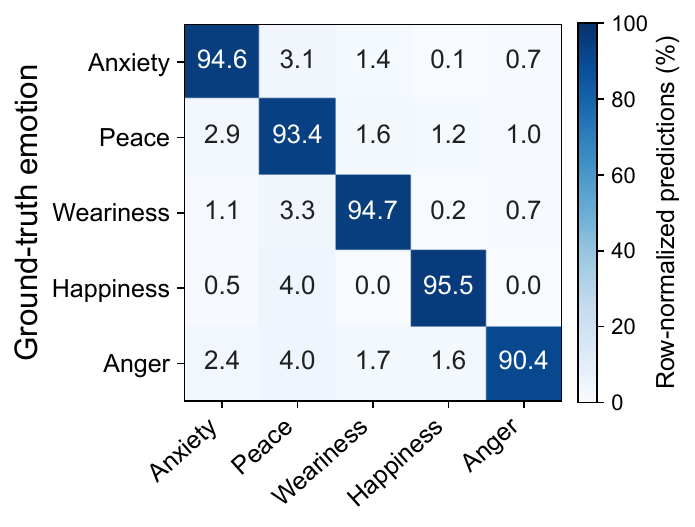}
        \caption{Fused output on AIDE}
        \label{fig:confuse_fusion_aide}
    \end{subfigure}

    \vspace{0.5em}

        \begin{subfigure}[t]{0.31\textwidth}
        \centering
        \includegraphics[width=\linewidth]{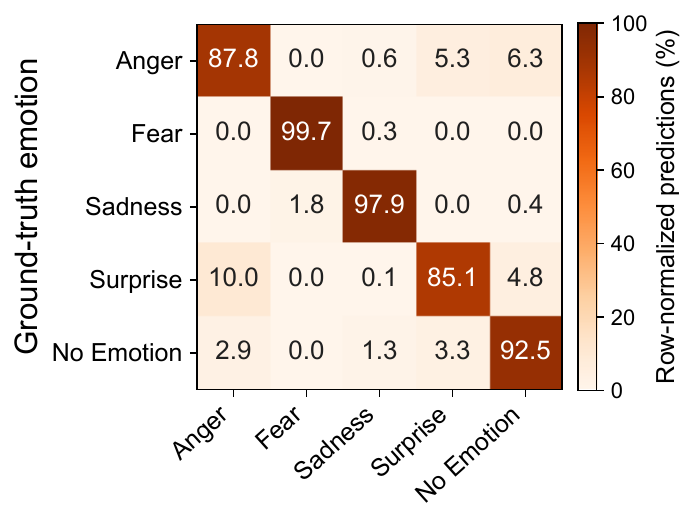}
        \caption{Visual branch on manD}
        \label{fig:confuse_face_mand}
    \end{subfigure}
    \hfill
    \begin{subfigure}[t]{0.31\textwidth}
        \centering
        \includegraphics[width=\linewidth]{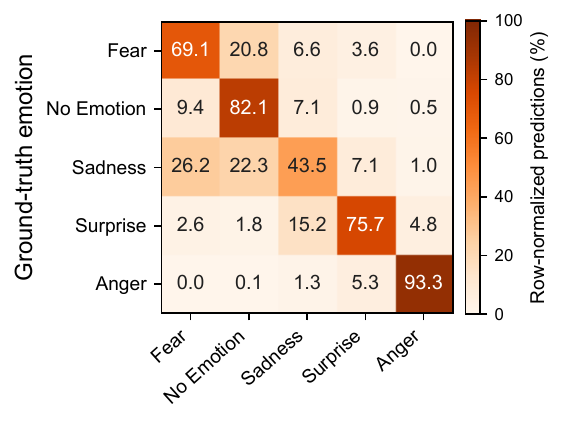}
        \caption{Traffic context branch on manD}
        \label{fig:confuse_context_mand}
    \end{subfigure}
    \hfill
    \begin{subfigure}[t]{0.31\textwidth}
        \centering
        \includegraphics[width=\linewidth]{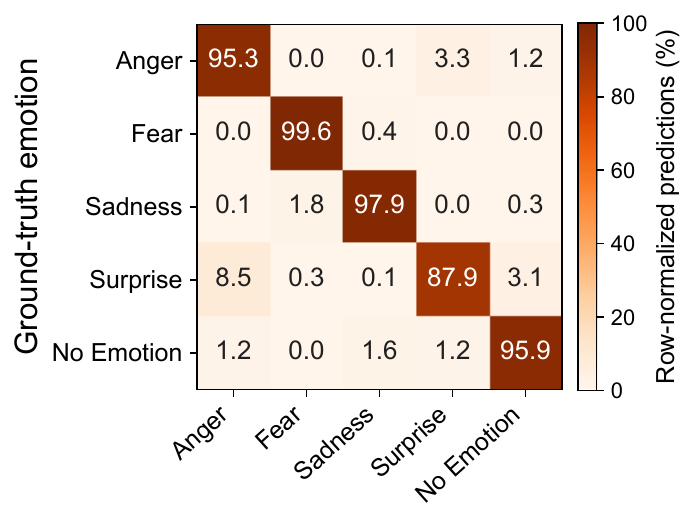}
        \caption{Fused output on manD}
        \label{fig:confuse_fusion_mand}
    \end{subfigure}

    \caption{Mean row-normalized confusion matrices over five random seeds on AIDE and manD.}
    \label{fig:confuse_matrix_modalities}
\end{figure*}

\begin{figure*}[htbp]
    \centering

        \begin{subfigure}[t]{0.31\textwidth}
        \centering
        \includegraphics[width=\linewidth]{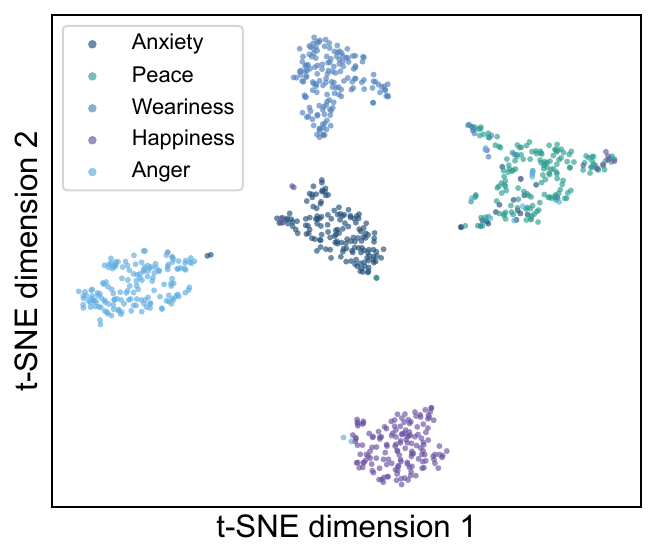}
        \caption{Visual outputs on AIDE}
        \label{fig:tsne_face_aide}
    \end{subfigure}
    \hfill
    \begin{subfigure}[t]{0.31\textwidth}
        \centering
        \includegraphics[width=\linewidth]{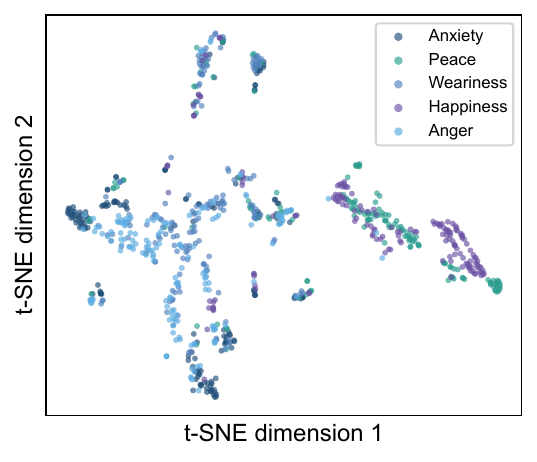}
        \caption{Traffic context outputs on AIDE}
        \label{fig:tsne_context_aide}
    \end{subfigure}
    \hfill
    \begin{subfigure}[t]{0.31\textwidth}
        \centering
        \includegraphics[width=\linewidth]{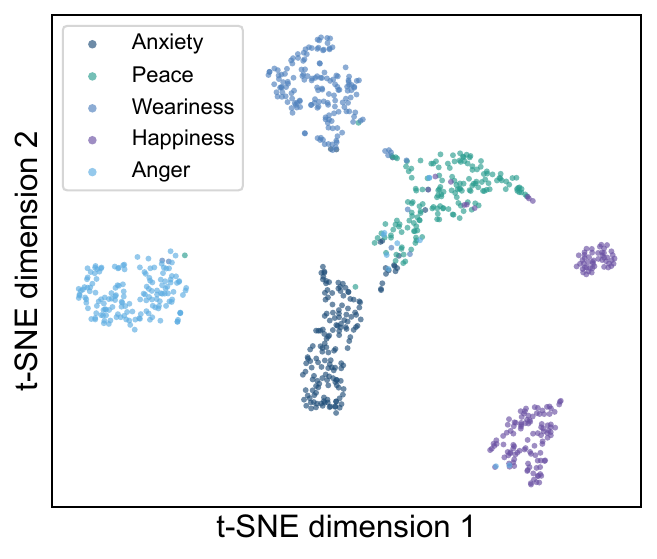}
        \caption{Fused outputs on AIDE}
        \label{fig:tsne_fusion_aide}
    \end{subfigure}

    \vspace{0.5em}

        \begin{subfigure}[t]{0.31\textwidth}
        \centering
        \includegraphics[width=\linewidth]{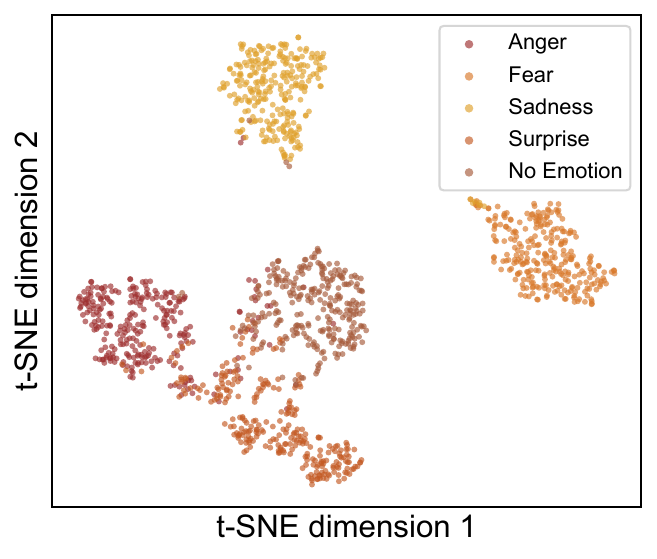}
        \caption{Visual outputs on manD}
        \label{fig:tsne_face_mand}
    \end{subfigure}
    \hfill
    \begin{subfigure}[t]{0.31\textwidth}
        \centering
        \includegraphics[width=\linewidth]{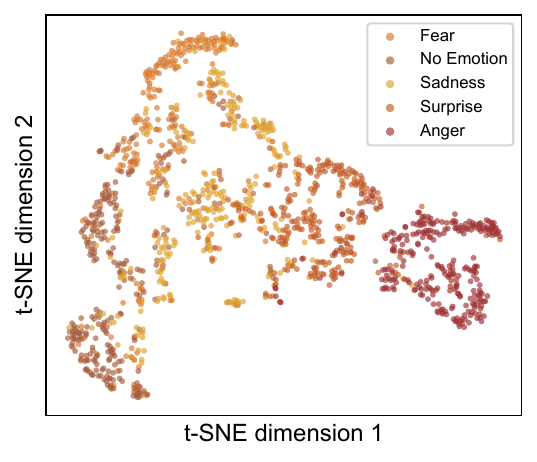}
        \caption{Traffic context outputs on manD}
        \label{fig:tsne_context_mand}
    \end{subfigure}
    \hfill
    \begin{subfigure}[t]{0.31\textwidth}
        \centering
        \includegraphics[width=\linewidth]{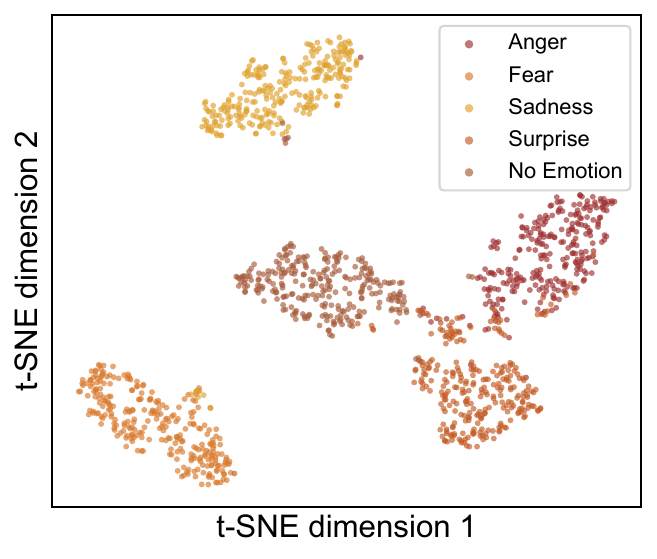}
        \caption{Fused outputs on manD}
        \label{fig:tsne_fusion_mand}
    \end{subfigure}

    \caption{Descriptive t-SNE projections of the standardized pre-softmax outputs averaged across five random seeds on AIDE and manD.}
    \label{fig:t-SNE_modalities}
\end{figure*}

In this subsection, we evaluate the emotion recognition performance of TADER on the AIDE and manD test sets.
We adopt class-wise recall, overall accuracy, and Macro-F1 as the primary evaluation metrics, with all results reported as the mean $\pm$ standard deviation over five runs using different random seeds.
For overall accuracy and Macro-F1, we report both fine-grained and coarse-grained evaluations. For AIDE, the coarse-grained categories are defined as Negative (\emph{Anxiety}, \emph{Weariness}, and \emph{Anger}), Neutral (\emph{Peace}), and Positive (\emph{Happiness}); for manD, they are defined as Negative (\emph{Fear}, \emph{Sadness}, \emph{Surprise}, and \emph{Anger}) and Neutral (\emph{No Emotion}). The coarse-grained predictions are obtained by mapping both the ground-truth and predicted fine-grained labels to these aggregated categories before recomputing accuracy and Macro-F1.
Table~\ref{tab:benchmark} summarizes the comparison with the visual-only benchmarks.
We further analyze the contributions of the visual and traffic context branches using confusion matrices in Fig.~\ref{fig:confuse_matrix_modalities} and t-SNE visualizations in Fig.~\ref{fig:t-SNE_modalities}.

As shown in Table~\ref{tab:benchmark}, the visual-only baselines exhibit substantial performance gaps. MTCNN-D yields the poorest results, with Macro-F1 scores of 67.26\% on AIDE and 53.42\% on manD. 
MAE and IAL improve substantially over this conventional baseline, while CAERNet achieves the highest fine-grained Macro-F1 among the visual-only baselines, with scores of 88.93\% and 90.01\% on AIDE and manD, respectively.

Against the matched CAERNet baseline, TADER consistently improves recognition performance on both datasets. On AIDE, Macro-F1 increases from 88.93\% to 93.75\%, while on manD it increases from 90.01\% to 95.30\%. 
Moreover, TADER improves both fine-grained and coarse-grained Macro-F1 over the corresponding visual-only baseline for all four evaluated visual backbones. The largest gains are observed with MTCNN-D, while improvements remain evident for the other three visual backbones.
This pattern indicates that traffic context can compensate for ambiguous visual representations while still providing complementary information when the visual representation is already strong. 
The coarse-grained results further indicate improved discrimination at broader affective levels, particularly on manD, where the evaluation mainly distinguishes emotional responses from \emph{No Emotion}. The fine-grained results remain the primary evidence for distinguishing individual emotion categories.

The class-wise results further show emotion-specific and dataset-dependent improvements. On AIDE, the largest improvement over CAERNet occurs for \emph{Anxiety}, whose recall increases from 83.42\% to 94.58\%. On manD, particularly large gains are observed for \emph{Anger} and \emph{No Emotion}. These results indicate that traffic context provides complementary information beyond visual appearance, although the magnitude of its contribution varies across emotion categories and datasets.

\begin{table*}[htbp]
\centering
\caption{Performance comparison of different benchmarks on AIDE and manD.}
\label{tab:benchmark}
\renewcommand{\arraystretch}{1.1}
\scriptsize
\setlength{\tabcolsep}{2pt}

\begin{subtable}{\linewidth}
\centering
\caption{Results on AIDE.}
\label{tab:benchmark_aide}
\begin{tabular}{l*{9}{c}}
\toprule
\multirow{2}{*}{Benchmark}
& \multicolumn{5}{c}{Recall of Each Emotion (\%)}
& \multicolumn{2}{c}{Fine-grained (\%)}
& \multicolumn{2}{c}{Coarse-grained (\%)} \\
\cmidrule(lr){2-6}\cmidrule(lr){7-8}\cmidrule(lr){9-10}
& Anxiety & Peace & Weariness & Happiness & Anger
& Acc & F1 & Acc & F1 \\
\midrule

MTCNN-D
& $37.23\pm6.73$ & $98.80\pm1.20$ & $73.98\pm8.43$
& $77.59\pm3.91$ & $55.66\pm2.88$
& $68.65\pm0.92$ & $67.26\pm0.50$
& $84.00\pm1.80$ & $82.45\pm1.65$ \\

MAE
& $72.78\pm12.80$ & $86.28\pm5.75$ & $85.55\pm2.33$
& $82.66\pm9.75$ & $77.00\pm16.49$
& $80.86\pm7.73$ & $80.82\pm7.90$
& $87.17\pm2.61$ & $85.52\pm3.17$ \\

IAL
& $74.30\pm7.25$ & $95.90\pm1.88$ & $83.90\pm4.48$
& $89.10\pm4.16$ & $90.40\pm7.06$
& $86.90\pm1.22$ & $86.90\pm1.57$
& $91.20\pm1.29$ & $90.00\pm0.89$ \\

CAERNet
& $83.42\pm1.83$ & $88.96\pm3.10$ & $92.46\pm1.67$
& $91.13\pm0.54$ & $88.48\pm5.54$
& $88.89\pm0.82$ & $88.93\pm0.81$
& $90.80\pm0.81$ & $89.63\pm0.85$ \\

TADER (MTCNN-D)
& $81.37\pm3.96$ & $92.10\pm2.67$ & $89.08\pm2.39$
& $91.86\pm2.06$ & $78.72\pm6.36$
& $86.63\pm0.97$ & $86.55\pm1.01$
& $92.53\pm0.58$ & $91.58\pm0.69$ \\

TADER (MAE)
& $85.76\pm4.00$ & $89.37\pm1.35$ & $89.49\pm2.10$
& $90.94\pm0.69$ & $90.46\pm2.31$
& $89.20\pm1.25$ & $89.26\pm1.28$
& $90.80\pm0.65$ & $89.88\pm0.74$ \\

TADER (IAL)
& $86.00\pm4.12$ & $95.90\pm2.13$ & $90.10\pm3.12$
& $88.20\pm9.22$ & $90.10\pm4.43$
& $90.10\pm2.17$ & $90.00\pm2.16$
& $94.30\pm0.67$ & $93.40\pm1.31$ \\

\textbf{TADER (CAERNet)}
& $94.58\pm2.59$ & $93.37\pm1.13$ & $94.70\pm1.67$
& $95.54\pm0.69$ & $90.36\pm1.90$
& $93.71\pm0.79$ & $93.75\pm0.79$
& $95.33\pm0.48$ & $94.34\pm0.53$ \\

\bottomrule
\end{tabular}
\end{subtable}

\vspace{1.0em}

\begin{subtable}{\linewidth}
\centering
\caption{Results on manD.}
\label{tab:benchmark_mand}
\begin{tabular}{l*{9}{c}}
\toprule
\multirow{2}{*}{Benchmark}
& \multicolumn{5}{c}{Recall of Each Emotion (\%)}
& \multicolumn{2}{c}{Fine-grained (\%)}
& \multicolumn{2}{c}{Coarse-grained (\%)} \\
\cmidrule(lr){2-6}\cmidrule(lr){7-8}\cmidrule(lr){9-10}
& Fear & No Emo. & Sadness & Surprise & Anger
& Acc & F1 & Acc & F1 \\
\midrule

MTCNN-D
& $99.13\pm1.55$ & $24.29\pm10.38$ & $81.54\pm6.20$
& $19.21\pm10.63$ & $57.27\pm7.81$
& $56.21\pm0.93$ & $53.42\pm2.79$
& $77.36\pm2.39$ & $57.73\pm2.72$ \\

MAE
& $67.10\pm1.23$ & $97.12\pm1.52$ & $92.62\pm2.10$
& $75.32\pm1.71$ & $93.67\pm0.89$
& $85.17\pm1.42$ & $86.02\pm1.28$
& $85.18\pm1.42$ & $81.14\pm1.63$ \\

IAL
& $97.25\pm0.55$ & $69.21\pm1.84$ & $98.99\pm0.33$
& $93.46\pm0.66$ & $72.10\pm1.45$
& $86.20\pm0.50$ & $85.72\pm0.49$
& $92.78\pm0.29$ & $87.47\pm0.55$ \\

CAERNet
& $97.86\pm0.20$ & $88.84\pm4.09$ & $94.69\pm1.56$
& $84.86\pm2.46$ & $83.90\pm3.47$
& $90.02\pm1.20$ & $90.01\pm1.18$
& $94.11\pm0.52$ & $91.03\pm0.81$ \\

TADER (MTCNN-D)
& $92.00\pm0.00$ & $82.25\pm3.88$ & $90.29\pm1.70$
& $77.39\pm5.85$ & $87.63\pm1.45$
& $85.89\pm0.66$ & $85.71\pm0.79$
& $93.25\pm0.57$ & $89.36\pm1.05$ \\

TADER (MAE)
& $99.20\pm0.47$ & $92.36\pm1.29$ & $95.22\pm2.48$
& $86.28\pm2.99$ & $88.99\pm3.14$
& $92.39\pm1.09$ & $92.41\pm1.09$
& $96.71\pm0.58$ & $94.88\pm0.84$ \\

TADER (IAL)
& $96.40\pm2.49$ & $90.60\pm3.40$ & $90.90\pm3.93$
& $83.80\pm3.08$ & $98.20\pm0.32$
& $92.00\pm0.97$ & $91.90\pm0.95$
& $97.10\pm0.55$ & $95.40\pm0.88$ \\

\textbf{TADER (CAERNet)}
& $99.57\pm0.47$ & $95.93\pm2.54$ & $97.87\pm0.60$
& $87.94\pm1.85$ & $95.32\pm1.11$
& $95.31\pm0.89$ & $95.30\pm0.89$
& $98.26\pm0.62$ & $97.28\pm0.96$ \\

\bottomrule
\end{tabular}
\end{subtable}
\end{table*}

\vspace{0.3em}
\noindent \emph{Value of multimodal fusion}.
We further examine how the visual and traffic context branches contribute to the final recognition using the confusion matrices in Fig.~\ref{fig:confuse_matrix_modalities} and the t-SNE visualizations in Fig.~\ref{fig:t-SNE_modalities}. The two branches exhibit clearly different recognition patterns on both datasets, showing that they provide different sources of information for driver emotion recognition. 
Notice that the visual outputs analyzed here are obtained from the visual branch within the jointly trained TADER framework and therefore differ from the independently trained CAERNet benchmark in Table~\ref{tab:benchmark}.

On AIDE, the visual branch already provides strong emotion-discriminative information, whereas the Traffic-only representation is substantially weaker when used independently. Nevertheless, the two representations exhibit different error patterns. The Traffic-only model retains meaningful discrimination for several categories, including \emph{Anxiety} and \emph{Peace}, while fusion further improves the visual representation, most notably increasing \emph{Anxiety} recall from 87.2\% to 94.6\%. This indicates that traffic context is not sufficient for fine-grained emotion recognition by itself, but provides complementary information that helps resolve visually ambiguous cases.

The complementarity is more evident on manD. Although the visual branch provides the stronger overall representation, the Traffic-only model shows substantial discriminative capability for specific categories, particularly \emph{Anger} (93.3\%), \emph{No Emotion} (82.1\%), and \emph{Surprise} (75.7\%). Notably, \emph{Anger} recall increases from 87.8\% in the visual branch to 95.3\% after fusion, while \emph{No Emotion} increases from 92.5\% to 95.9\%. These results show that the traffic context representation captures class-dependent information that differs from facial appearance, allowing the fused model to exploit complementary evidence from the two modalities.

The t-SNE visualizations in Fig.~\ref{fig:t-SNE_modalities} further illustrate this complementarity. The visual outputs already show relatively clear separation among emotion categories on both datasets, whereas the traffic context outputs exhibit much greater overlap, particularly on manD. This reflects the different roles of the two branches: the visual branch provides direct evidence of the driver's affective state, while the traffic context branch captures additional information from multi-vehicle motion and expectation deviation. After fusion, the output distributions become more compact and more clearly separated. This improvement is consistent with the confusion-matrix gains for \emph{Anxiety} and \emph{Anger} on AIDE and for \emph{Anger} and \emph{No Emotion} on manD. The results show that traffic context information complements the visual representation by providing additional information for visually ambiguous cases, leading to more reliable final emotion recognition.

\subsection{Ablation Study}
\label{sec:ablation}

In this subsection, we conduct three complementary ablation studies to examine the traffic context representation in TADER. First, we remove the actual trajectories $\sigma$, expectation deviations $d$, and prediction confidence $\rho$ individually to quantify their conditional contributions when the other two components are retained. Second, we vary the represented vehicle scope among the ego vehicle, surrounding vehicles, and their joint multi-vehicle representation to determine whether either source alone is sufficient. Third, we evaluate the contribution of transfer learning for adapting the trajectory predictor to the target emotion-recognition datasets.

\vspace{0.3em}
\noindent \emph{Value of traffic context representation components}.
The traffic context representation in TADER integrates three complementary components: the actual multi-vehicle trajectories $\sigma$, the expectation deviations $d$, and the prediction confidence $\rho$. To isolate their contributions, we remove each component individually while retaining all other settings the same.

For the traffic context branch alone, removing any component reduces Fine-grained Macro-F1 on both datasets. On AIDE, Macro-F1 decreases from 60.37\% to 59.93\%, 53.28\%, and 52.39\% when $\rho$, $d$, and $\sigma$ are removed, respectively. On manD, the corresponding values decrease from 72.22\% to 71.99\%, 66.49\%, and 57.33\%. The largest degradation is consistently caused by removing $\sigma$, followed by $d$, while $\rho$ provides a smaller but positive contribution.

The same aggregate ordering is observed in the complete TADER model. On AIDE, removing $\rho$, $d$, and $\sigma$ reduces Macro-F1 from 93.75\% to 93.66\%, 92.26\%, and 91.67\%, respectively. On manD, Macro-F1 decreases from 95.30\% to 94.71\%, 94.44\%, and 93.75\%. 
These results indicate that all three components contribute positively to the mean fine-grained Macro-F1 of the TADER model. Notably, the degradation caused by removing individual traffic context components is substantially larger in the Traffic-only model than in the complete TADER model, indicating that the visual representation can partially compensate for missing traffic context information. 

The class-wise results further show that the contribution of each component varies across emotions. Removing an individual component can improve the recall of particular categories while reducing aggregate Accuracy and Macro-F1. 
Overall, $\sigma$ provides the realized motion evidence, $d$ provides expectation-relative information, and $\rho$ supplies covariance-normalized prediction confidence for interpreting the agreement between observed and predicted trajectories.

\begin{table*}[htbp]
\centering
\caption{Ablation of the traffic context representation components on AIDE and manD.}
\label{tab:trajectory_scope}
\renewcommand{\arraystretch}{1.1}
\scriptsize
\setlength{\tabcolsep}{2pt}

\begin{subtable}{\linewidth}
\centering
\caption{Results on AIDE.}
\label{tab:trajectory_scope_aide}
\begin{tabular}{l*{9}{c}}
\toprule
\multirow{2}{*}{Variant}
& \multicolumn{5}{c}{Recall of Each Emotion (\%)}
& \multicolumn{2}{c}{Fine-grained (\%)}
& \multicolumn{2}{c}{Coarse-grained (\%)} \\
\cmidrule(lr){2-6}\cmidrule(lr){7-8}\cmidrule(lr){9-10}
& Anxiety & Peace & Weariness & Happiness & Anger
& Acc & F1 & Acc & F1 \\
\midrule

TADER (Traffic-only w/o $\sigma$)
& $67.83\pm3.65$ & $67.23\pm13.45$ & $39.28\pm3.02$
& $49.04\pm1.63$ & $40.24\pm6.51$
& $52.72\pm2.97$ & $52.39\pm3.05$
& $70.84\pm2.63$ & $64.11\pm3.78$ \\

TADER (Traffic-only w/o $d$)
& $70.36\pm1.37$ & $46.75\pm0.91$ & $36.51\pm7.30$
& $52.41\pm1.20$ & $62.41\pm3.53$
& $53.69\pm1.38$ & $53.28\pm1.56$
& $72.84\pm0.73$ & $63.20\pm0.84$ \\

TADER (Traffic-only w/o $\rho$)
& $70.60\pm1.16$ & $68.92\pm7.20$ & $53.13\pm5.89$
& $51.93\pm5.14$ & $54.58\pm4.03$
& $59.83\pm2.98$ & $59.93\pm3.07$
& $77.25\pm3.90$ & $70.32\pm4.72$ \\

TADER (Traffic-only)
& $70.60\pm2.06$ & $66.27\pm10.49$ & $56.14\pm12.41$
& $52.77\pm3.94$ & $55.78\pm4.40$
& $60.31\pm3.97$ & $60.37\pm4.24$
& $78.48\pm3.26$ & $71.20\pm4.83$ \\

TADER (Full w/o $\sigma$)
& $88.76\pm2.73$ & $91.77\pm3.30$ & $93.34\pm1.89$
& $93.10\pm1.31$ & $91.17\pm3.61$
& $91.63\pm0.37$ & $91.67\pm0.34$
& $93.46\pm0.27$ & $92.37\pm0.28$ \\

TADER (Full w/o $d$)
& $90.24\pm1.56$ & $90.84\pm2.96$ & $94.70\pm1.56$
& $93.61\pm0.91$ & $91.69\pm5.41$
& $92.22\pm0.98$ & $92.26\pm0.97$
& $93.86\pm0.61$ & $92.65\pm0.67$ \\

TADER (Full w/o $\rho$)
& $91.33\pm1.63$ & $94.10\pm1.44$ & $95.18\pm2.13$
& $94.70\pm1.44$ & $92.77\pm4.28$
& $93.61\pm1.03$ & $93.66\pm1.01$
& $95.30\pm0.83$ & $94.45\pm0.90$ \\

\textbf{TADER (Full)}
& $94.58\pm2.59$ & $93.37\pm1.13$ & $94.70\pm1.67$
& $95.54\pm0.69$ & $90.36\pm1.90$
& $93.71\pm0.79$ & $93.75\pm0.79$
& $95.33\pm0.48$ & $94.34\pm0.53$ \\

\bottomrule
\end{tabular}
\end{subtable}

\vspace{1.0em}

\begin{subtable}{\linewidth}
\centering
\caption{Results on manD.}
\label{tab:trajectory_scope_mand}
\begin{tabular}{l*{9}{c}}
\toprule
\multirow{2}{*}{Variant}
& \multicolumn{5}{c}{Recall of Each Emotion (\%)}
& \multicolumn{2}{c}{Fine-grained (\%)}
& \multicolumn{2}{c}{Coarse-grained (\%)} \\
\cmidrule(lr){2-6}\cmidrule(lr){7-8}\cmidrule(lr){9-10}
& Fear & No Emo. & Sadness & Surprise & Anger
& Acc & F1 & Acc & F1 \\
\midrule

TADER (Traffic-only w/o $\sigma$)
& $53.48\pm4.82$ & $77.67\pm2.07$ & $16.18\pm9.72$
& $59.13\pm5.41$ & $91.73\pm0.44$
& $59.80\pm2.20$ & $57.33\pm3.28$
& $80.35\pm1.19$ & $74.03\pm1.28$ \\

TADER (Traffic-only w/o $d$)
& $62.10\pm2.77$ & $83.05\pm0.80$ & $30.15\pm12.35$
& $69.39\pm5.74$ & $93.38\pm1.36$
& $67.75\pm2.39$ & $66.49\pm3.04$
& $83.92\pm2.24$ & $78.38\pm2.29$ \\

TADER (Traffic-only w/o $\rho$)
& $69.35\pm2.03$ & $81.96\pm2.19$ & $43.97\pm6.48$
& $74.08\pm3.42$ & $92.59\pm1.77$
& $72.50\pm1.36$ & $71.99\pm1.47$
& $87.43\pm0.79$ & $82.06\pm0.88$ \\

TADER (Traffic-only)
& $69.06\pm1.39$ & $82.11\pm1.07$ & $43.46\pm6.48$
& $75.67\pm3.15$ & $93.31\pm1.74$
& $72.83\pm1.06$ & $72.22\pm1.24$
& $87.49\pm0.66$ & $82.15\pm0.70$ \\

TADER (Full w/o $\sigma$)
& $98.42\pm0.18$ & $94.51\pm1.30$ & $94.18\pm2.15$
& $88.11\pm1.85$ & $93.97\pm1.60$
& $93.75\pm0.73$ & $93.75\pm0.73$
& $96.40\pm0.41$ & $94.52\pm0.60$ \\

TADER (Full w/o $d$)
& $97.82\pm0.24$ & $93.20\pm2.11$ & $95.13\pm0.95$
& $90.06\pm0.92$ & $95.84\pm0.67$
& $94.44\pm0.41$ & $94.44\pm0.41$
& $96.55\pm0.26$ & $94.68\pm0.42$ \\

TADER (Full w/o $\rho$)
& $98.26\pm0.31$ & $93.56\pm2.03$ & $95.28\pm1.03$
& $90.78\pm1.94$ & $95.91\pm1.06$
& $94.71\pm0.65$ & $94.71\pm0.65$
& $96.72\pm0.40$ & $94.94\pm0.64$ \\

\textbf{TADER (Full)}
& $99.57\pm0.47$ & $95.93\pm2.54$ & $97.87\pm0.60$
& $87.94\pm1.85$ & $95.32\pm1.11$
& $95.31\pm0.89$ & $95.30\pm0.89$
& $98.26\pm0.62$ & $97.28\pm0.96$ \\

\bottomrule
\end{tabular}
\end{subtable}
\end{table*}

\vspace{0.3em}
\noindent \emph{Value of traffic context scope}.
We next investigate the value of traffic context scope by comparing ego-only, surrounding-only, and full multi-vehicle representations. The ego-only and surrounding-only variants retain only the corresponding vehicle scope, while the full model jointly represents both the ego vehicle and surrounding vehicles under the same interaction-conditioned trajectory prediction framework.

As reported in Table~\ref{tab:contexts}, restricting the represented vehicle scope reduces aggregate performance on both datasets. On AIDE, the ego-only and surrounding-only variants achieve Macro-F1 scores of 91.23\% and 91.77\%, respectively, whereas the full model reaches 93.75\%. On manD, the corresponding values are 93.83\%, 92.89\%, and 95.30\%. The relative performance of the two restricted variants reverses across datasets: surrounding-only performs slightly better on AIDE, whereas ego-only performs better on manD. This reversal suggests that there is no universally dominant source of traffic context information. Instead, the relative value of ego and surrounding-vehicle motion depends on the interaction patterns represented in the dataset. Nevertheless, their joint representation consistently achieves the highest aggregate performance, supporting the use of multi-vehicle context rather than an ego-centric or surrounding-only formulation.

\begin{table*}[htbp]
\centering
\caption{Ablation of the represented vehicle scope on AIDE and manD.}
\label{tab:contexts}
\renewcommand{\arraystretch}{1.1}
\scriptsize
\setlength{\tabcolsep}{2pt}

\begin{subtable}{\linewidth}
\centering
\caption{Results on AIDE.}
\label{tab:contexts_aide}
\begin{tabular}{l*{9}{c}}
\toprule
\multirow{2}{*}{Variant}
& \multicolumn{5}{c}{Recall of Each Emotion (\%)}
& \multicolumn{2}{c}{Fine-grained (\%)}
& \multicolumn{2}{c}{Coarse-grained (\%)} \\
\cmidrule(lr){2-6}\cmidrule(lr){7-8}\cmidrule(lr){9-10}
& Anxiety & Peace & Weariness & Happiness & Anger
& Acc & F1 & Acc & F1 \\
\midrule

TADER (Ego only)
& $89.69\pm2.42$ & $92.22\pm1.84$ & $93.66\pm1.08$
& $93.54\pm1.25$ & $87.04\pm5.46$
& $91.23\pm1.03$ & $91.23\pm1.07$
& $93.11\pm0.52$ & $92.04\pm0.49$ \\

TADER (Surrounding only)
& $89.04\pm1.31$ & $92.53\pm1.01$ & $94.94\pm1.09$
& $92.77\pm1.48$ & $89.16\pm3.19$
& $91.69\pm0.85$ & $91.77\pm0.86$
& $92.94\pm0.57$ & $91.80\pm0.55$ \\

\textbf{TADER (Full)}
& $94.58\pm2.59$ & $93.37\pm1.13$ & $94.70\pm1.67$
& $95.54\pm0.69$ & $90.36\pm1.90$
& $93.71\pm0.79$ & $93.75\pm0.79$
& $95.33\pm0.48$ & $94.34\pm0.53$ \\

\bottomrule
\end{tabular}
\end{subtable}

\vspace{1.0em}

\begin{subtable}{\linewidth}
\centering
\caption{Results on manD.}
\label{tab:contexts_mand}
\begin{tabular}{l*{9}{c}}
\toprule
\multirow{2}{*}{Variant}
& \multicolumn{5}{c}{Recall of Each Emotion (\%)}
& \multicolumn{2}{c}{Fine-grained (\%)}
& \multicolumn{2}{c}{Coarse-grained (\%)} \\
\cmidrule(lr){2-6}\cmidrule(lr){7-8}\cmidrule(lr){9-10}
& Fear & No Emo. & Sadness & Surprise & Anger
& Acc & F1 & Acc & F1 \\
\midrule

TADER (Ego only)
& $96.71\pm0.30$ & $94.53\pm1.13$ & $95.01\pm0.42$
& $90.21\pm2.03$ & $92.76\pm0.69$
& $93.84\pm0.47$ & $93.83\pm0.47$
& $95.61\pm0.35$ & $93.41\pm0.52$ \\

TADER (Surrounding only)
& $99.49\pm0.41$ & $92.15\pm2.50$ & $97.50\pm0.60$
& $87.15\pm2.16$ & $88.20\pm2.06$
& $92.87\pm0.67$ & $92.89\pm0.66$
& $96.57\pm0.20$ & $94.66\pm0.39$ \\

\textbf{TADER (Full)}
& $99.57\pm0.47$ & $95.93\pm2.54$ & $97.87\pm0.60$
& $87.94\pm1.85$ & $95.32\pm1.11$
& $95.31\pm0.89$ & $95.30\pm0.89$
& $98.26\pm0.62$ & $97.28\pm0.96$ \\

\bottomrule
\end{tabular}
\end{subtable}
\end{table*}

\vspace{0.3em}
\noindent \emph{Value of transfer learning}.
Table~\ref{tab:transfer_ablation} evaluates the contribution of transfer learning for the trajectory predictor. On AIDE, transfer learning increases Fine-grained Macro-F1 from 92.20\% to 93.75\% and coarse-grained Macro-F1 from 92.64\% to 94.34\%. On manD, the corresponding values increase from 94.16\% to 95.30\% and from 94.91\% to 97.28\%, respectively. These results indicate that transferring trajectory knowledge improves the aggregate emotion-recognition performance, although the class-wise effects remain emotion-dependent.

\begin{table*}[htbp]
\centering
\caption{Recognition accuracy of transfer learning ablations on AIDE and manD.}
\label{tab:transfer_ablation}
\renewcommand{\arraystretch}{1.1}
\scriptsize
\setlength{\tabcolsep}{2pt}

\begin{subtable}{\linewidth}
\centering
\caption{Results on AIDE.}
\label{tab:transfer_ablation_aide}
\begin{tabular}{l*{9}{c}}
\toprule
\multirow{2}{*}{Variant}
& \multicolumn{5}{c}{Recall of Each Emotion (\%)}
& \multicolumn{2}{c}{Fine-grained (\%)}
& \multicolumn{2}{c}{Coarse-grained (\%)} \\
\cmidrule(lr){2-6}\cmidrule(lr){7-8}\cmidrule(lr){9-10}
& Anxiety & Peace & Weariness & Happiness & Anger
& Acc & F1 & Acc & F1 \\
\midrule

TADER (Full w/o transfer)
& $89.45\pm2.99$ & $93.30\pm2.42$ & $93.54\pm1.63$
& $93.54\pm0.54$ & $90.89\pm1.62$
& $92.14\pm0.66$ & $92.20\pm0.68$
& $93.54\pm0.24$ & $92.64\pm0.37$ \\

\textbf{TADER (Full w/ transfer)}
& $94.58\pm2.59$ & $93.37\pm1.13$ & $94.70\pm1.67$
& $95.54\pm0.69$ & $90.36\pm1.90$
& $93.71\pm0.79$ & $93.75\pm0.79$
& $95.33\pm0.48$ & $94.34\pm0.53$ \\

\bottomrule
\end{tabular}
\end{subtable}

\vspace{1.0em}

\begin{subtable}{\linewidth}
\centering
\caption{Results on manD.}
\label{tab:transfer_ablation_mand}
\begin{tabular}{l*{9}{c}}
\toprule
\multirow{2}{*}{Variant}
& \multicolumn{5}{c}{Recall of Each Emotion (\%)}
& \multicolumn{2}{c}{Fine-grained (\%)}
& \multicolumn{2}{c}{Coarse-grained (\%)} \\
\cmidrule(lr){2-6}\cmidrule(lr){7-8}\cmidrule(lr){9-10}
& Fear & No Emo. & Sadness & Surprise & Anger
& Acc & F1 & Acc & F1 \\
\midrule

TADER (Full w/o transfer)
& $98.57\pm0.47$ & $94.20\pm2.38$ & $97.31\pm0.42$
& $88.53\pm1.89$ & $92.31\pm2.78$
& $94.17\pm0.88$ & $94.16\pm0.88$
& $96.68\pm0.21$ & $94.91\pm0.36$ \\

\textbf{TADER (Full w/ transfer)}
& $99.57\pm0.47$ & $95.93\pm2.54$ & $97.87\pm0.60$
& $87.94\pm1.85$ & $95.32\pm1.11$
& $95.31\pm0.89$ & $95.30\pm0.89$
& $98.26\pm0.62$ & $97.28\pm0.96$ \\

\bottomrule
\end{tabular}
\end{subtable}
\end{table*}

\subsection{Sensitivity Analysis}
\label{sec:sensitivity}

In this subsection, we examine the sensitivity of TADER to two implementation choices: the source dataset used to pretrain the interaction-conditioned trajectory predictor and the trajectory predictor architecture. These analyses determine whether the observed gains depend on a specific source-domain trajectory prior or prediction model.

\vspace{0.3em}
\noindent \emph{Sensitivity analysis on the trajectory-pretraining source}.
We first investigate whether TADER is sensitive to the dataset used to pretrain its interaction-conditioned trajectory predictor. Specifically, the predictor is pretrained using either NGSIM or highD before being integrated into the complete recognition framework. 
As shown in Table~\ref{tab:source_sensitivity}, both pretraining sources support strong performance on AIDE. Using NGSIM produces an accuracy of $93.71\pm0.79\%$ and a Macro-F1 of $93.75\pm0.79\%$, while using highD increases these values to $94.39\pm0.97\%$ and $94.43\pm0.98\%$, respectively. The resulting Macro-F1 difference is 0.68 points, indicating that the overall performance of TADER is not strongly dependent on a particular trajectory-pretraining corpus.

When focusing on class-wise changes, we find that relative to NGSIM pretraining, highD pretraining improves the recalls of \emph{Peace}, \emph{Weariness}, and \emph{Anger} by 1.21, 0.24, and 4.22 points, respectively, while reducing \emph{Anxiety} and \emph{Happiness} by 1.93 and 0.36 points. Thus, changing the pretraining source does not uniformly improve all categories. Instead, it alters the motion prior learned by the trajectory predictor and consequently redistributes the contribution of traffic context across different emotions. In particular, the substantial increase for \emph{Anger} and the decrease for \emph{Anxiety} suggest that these categories benefit from different aspects of the pretrained motion representation.
A similar pattern is observed on manD. HighD pretraining increases accuracy from $95.31\pm0.89\%$ to $96.28\pm0.34\%$ and Macro-F1 from $95.30\pm0.89\%$ to $96.28\pm0.35\%$, corresponding to improvements of 0.97 and 0.98 points, respectively. The largest class-wise gains occur for \emph{Surprise} and \emph{No Emotion}, whose recalls increase by 3.54 and 1.82 points. By contrast, \emph{Anger} and \emph{Fear} decrease slightly by 0.28 and 0.22 points, while \emph{Sadness} remains unchanged. 
The class-wise shifts across AIDE and manD suggest that different pretraining sources emphasize different motion patterns, causing the traffic context benefit to vary across emotions. However, the consistently strong aggregate performance indicates that TADER can effectively exploit trajectory knowledge learned from either source.

\begin{table*}[htbp]
\centering
\caption{Sensitivity of TADER to the trajectory-predictor pretraining source on AIDE and manD.}
\label{tab:source_sensitivity}
\renewcommand{\arraystretch}{1.1}
\scriptsize
\setlength{\tabcolsep}{2pt}

\begin{subtable}{\linewidth}
\centering
\caption{Results on AIDE.}
\label{tab:source_sensitivity_aide}
\begin{tabular}{l*{9}{c}}
\toprule
\multirow{2}{*}{Pretraining Source}
& \multicolumn{5}{c}{Recall of Each Emotion (\%)}
& \multicolumn{2}{c}{Fine-grained (\%)}
& \multicolumn{2}{c}{Coarse-grained (\%)} \\
\cmidrule(lr){2-6}\cmidrule(lr){7-8}\cmidrule(lr){9-10}
& Anxiety & Peace & Weariness & Happiness & Anger
& Acc & F1 & Acc & F1 \\
\midrule
TADER (Full, NGSIM)
& $94.58\pm2.59$ & $93.37\pm1.13$ & $94.70\pm1.67$
& $95.54\pm0.69$ & $90.36\pm1.90$
& $93.71\pm0.79$ & $93.75\pm0.79$
& $95.33\pm0.48$ & $94.34\pm0.53$ \\

TADER (Full, highD)
& $92.65\pm2.15$ & $94.58\pm1.59$ & $94.94\pm1.09$
& $95.18\pm1.76$ & $94.58\pm3.10$
& $94.39\pm0.97$ & $94.43\pm0.98$
& $95.88\pm0.51$ & $94.92\pm0.55$ \\
\bottomrule
\end{tabular}
\end{subtable}

\vspace{1.0em}

\begin{subtable}{\linewidth}
\centering
\caption{Results on manD.}
\label{tab:source_sensitivity_mand}
\begin{tabular}{l*{9}{c}}
\toprule
\multirow{2}{*}{Pretraining Source}
& \multicolumn{5}{c}{Recall of Each Emotion (\%)}
& \multicolumn{2}{c}{Fine-grained (\%)}
& \multicolumn{2}{c}{Coarse-grained (\%)} \\
\cmidrule(lr){2-6}\cmidrule(lr){7-8}\cmidrule(lr){9-10}
& Fear & No Emo. & Sadness & Surprise & Anger
& Acc & F1 & Acc & F1 \\
\midrule
TADER (Full, NGSIM)
& $99.57\pm0.47$ & $95.93\pm2.54$ & $97.87\pm0.60$
& $87.94\pm1.85$ & $95.32\pm1.11$
& $95.31\pm0.89$ & $95.30\pm0.89$
& $98.26\pm0.62$ & $97.28\pm0.96$ \\

TADER (Full, highD)
& $99.35\pm0.40$ & $97.75\pm0.70$ & $97.87\pm0.31$
& $91.48\pm2.07$ & $95.04\pm1.36$
& $96.28\pm0.34$ & $96.28\pm0.35$
& $98.78\pm0.22$ & $98.10\pm0.33$ \\
\bottomrule
\end{tabular}
\end{subtable}
\end{table*}

\vspace{0.3em}
\noindent \emph{Sensitivity analysis on the trajectory predictor architecture}.
We further examine whether TADER depends on a particular trajectory predictor architecture by comparing Pishgu with Trajectron++. As shown in Table~\ref{tab:predictor_architecture}, replacing Pishgu with Trajectron++ increases Fine-grained Macro-F1 from 93.75\% to 94.63\% on AIDE and from 95.30\% to 97.49\% on manD. The corresponding coarse-grained Macro-F1 increases from 94.34\% to 94.86\% and from 97.28\% to 98.93\%, respectively. These results indicate that TADER is not tied to a specific trajectory predictor and can benefit from stronger trajectory prediction models.

\begin{table*}[htbp]
\centering
\caption{Recognition accuracy with different trajectory predictor architectures on AIDE and manD.}
\label{tab:predictor_architecture}
\renewcommand{\arraystretch}{1.1}
\scriptsize
\setlength{\tabcolsep}{2pt}

\textbf{AIDE}

\vspace{0.3em}

\begin{tabular}{l*{9}{c}}
\toprule
\multirow{2}{*}{Variant}
& \multicolumn{5}{c}{Recall of Each Emotion (\%)}
& \multicolumn{2}{c}{Fine-grained (\%)}
& \multicolumn{2}{c}{Coarse-grained (\%)} \\
\cmidrule(lr){2-6}\cmidrule(lr){7-8}\cmidrule(lr){9-10}
& Anxiety & Peace & Weariness & Happiness & Anger
& Acc & F1 & Acc & F1 \\
\midrule

TADER (Full, Pishgu)
& $94.58\pm2.59$ & $93.37\pm1.13$ & $94.70\pm1.67$
& $95.54\pm0.69$ & $90.36\pm1.90$
& $93.71\pm0.79$ & $93.75\pm0.79$
& $95.33\pm0.48$ & $94.34\pm0.53$ \\

TADER (Full, Trajectron++)
& $93.01\pm1.57$ & $95.30\pm0.79$ & $94.82\pm1.39$
& $95.30\pm0.50$ & $94.46\pm2.23$
& $94.58\pm0.47$ & $94.63\pm0.46$
& $95.83\pm0.46$ & $94.86\pm0.43$ \\
\bottomrule
\end{tabular}

\vspace{1.0em}

\textbf{manD}

\vspace{0.3em}

\begin{tabular}{l*{9}{c}}
\toprule
\multirow{2}{*}{Variant}
& \multicolumn{5}{c}{Recall of Each Emotion (\%)}
& \multicolumn{2}{c}{Fine-grained (\%)}
& \multicolumn{2}{c}{Coarse-grained (\%)} \\
\cmidrule(lr){2-6}\cmidrule(lr){7-8}\cmidrule(lr){9-10}
& Fear & No Emo. & Sadness & Surprise & Anger
& Acc & F1 & Acc & F1 \\
\midrule

TADER (Full, Pishgu)
& $99.57\pm0.47$ & $95.93\pm2.54$ & $97.87\pm0.60$
& $87.94\pm1.85$ & $95.32\pm1.11$
& $95.31\pm0.89$ & $95.30\pm0.89$
& $98.26\pm0.62$ & $97.28\pm0.96$ \\

TADER (Full, Trajectron++)
& $99.42\pm0.71$ & $97.67\pm0.20$ & $98.09\pm0.31$
& $94.73\pm1.10$ & $97.55\pm0.86$
& $97.49\pm0.22$ & $97.49\pm0.22$
& $99.32\pm0.04$ & $98.93\pm0.06$ \\
\bottomrule
\end{tabular}

\end{table*}

\section{Conclusion}
\label{sec:Conclusion}

In this paper, we propose TADER, a traffic-context-aware driver emotion recognition framework that integrates in-cabin visual signals with out-of-cabin multi-vehicle traffic information. The proposed framework comprises two main components: (1) a vision-based recognizer that extracts spatiotemporal representations from in-cabin facial images to infer driver emotional states, and (2) a traffic context module that establishes interaction-conditioned motion references through probabilistic trajectory prediction and jointly represents realized trajectories $\sigma$, expectation deviations $d$, and prediction confidence $\rho$. The two sources of information are integrated to provide a more complete representation of driver emotion under ongoing traffic interactions. To address the limited availability of datasets with paired in-cabin visual and external trajectory data, we further introduce a transfer learning strategy that first pretrains the trajectory predictor on large-scale public trajectory data and then transfers the learned interaction knowledge to the target emotion-recognition datasets. 
Extensive experiments on AIDE and manD show that TADER consistently outperforms state-of-the-art methods across both naturalistic and simulated driving environments, resulting in improvements of 4.82\% in accuracy on the AIDE dataset, and improvements of 5.29\% in accuracy on the manD dataset.

This work opens several directions for future research. 
First, we plan to enrich the driving context with additional information, including pedestrian behavior, road topology, and weather conditions, and apply causal inference to better understand the contribution of different contextual factors to driver emotional responses.
Second, we plan to develop continual and personalized vehicle control strategies that leverage real-time emotion recognition to adapt vehicle behavior to the driver's affective state and preferences.

\bibliographystyle{IEEEtran}
\bibliography{References.bib}

\begin{IEEEbiography}
[{\includegraphics[width=1in,height=1.25in,clip,keepaspectratio]
{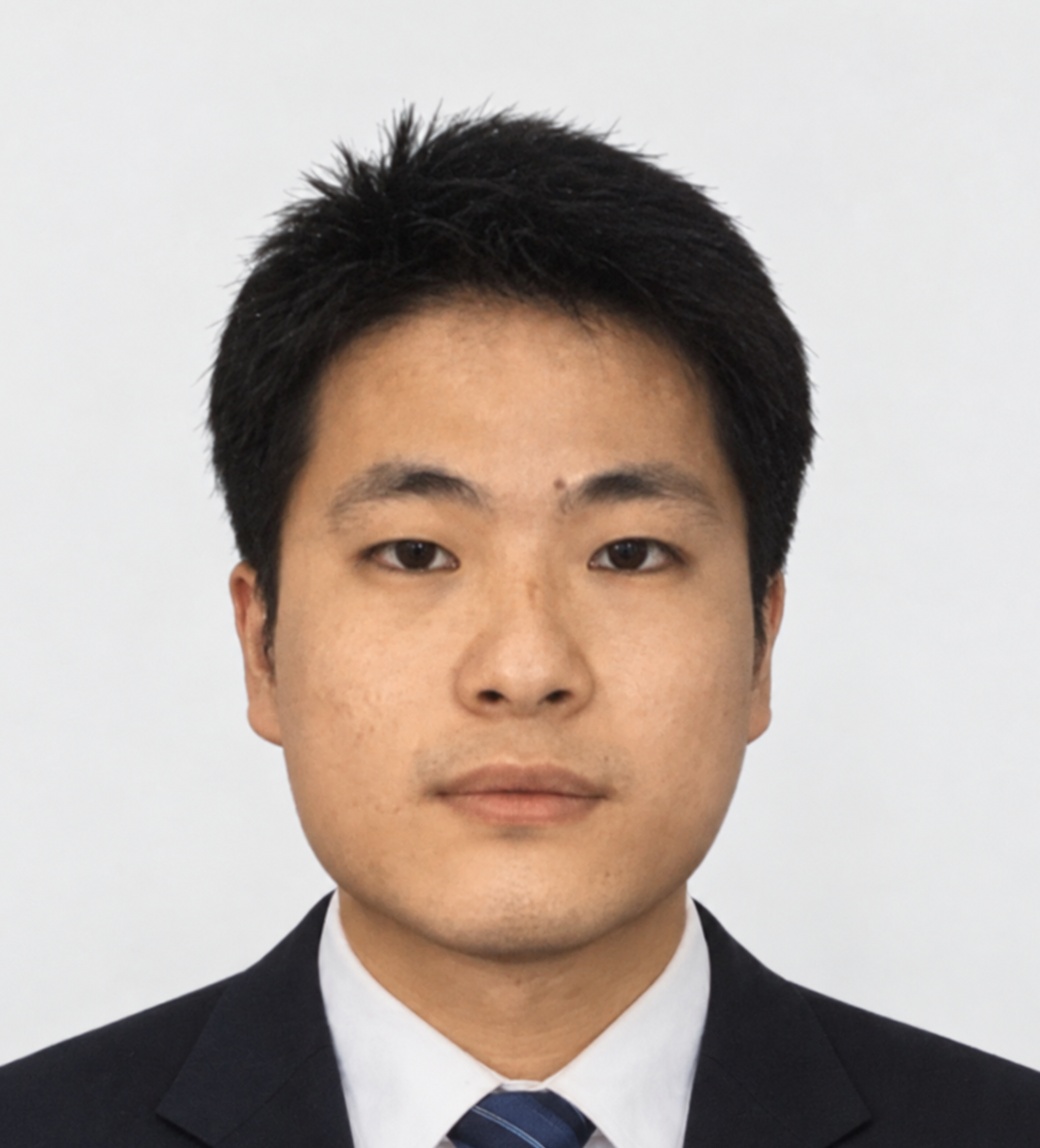}}]
{Longhao Yan} received the B.Eng. degree and M.Eng. degree in School of Electronics and Control Engineering from Chang’an University, Xi’an, China, in 2019 and 2022 respectively. He is currently working towards a Ph.D. degree with the National University of Singapore. His research interests include vehicle control and trajectory prediction of intelligent transportation system.
\end{IEEEbiography}
\begin{IEEEbiography}
[{\includegraphics[width=1in,height=1.25in,clip,keepaspectratio]{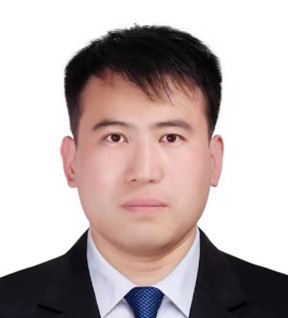}}]{Jinhao Liang} received the B.S. degree from School of Mechanical Engineering, Nanjing University of Science and Technology, Nanjing, China, in 2017, and Ph.D. degree from School of Mechanical Engineering, Southeast University, Nanjing, China, in 2022. Now he is a Research Fellow with Department of Civil and Environmental Engineering, National University of Singapore. His research interests include vehicle dynamics and control, and autonomous vehicles.
\end{IEEEbiography}
\begin{IEEEbiography}
[{\includegraphics[width=1in,height=1.25in,clip,keepaspectratio]{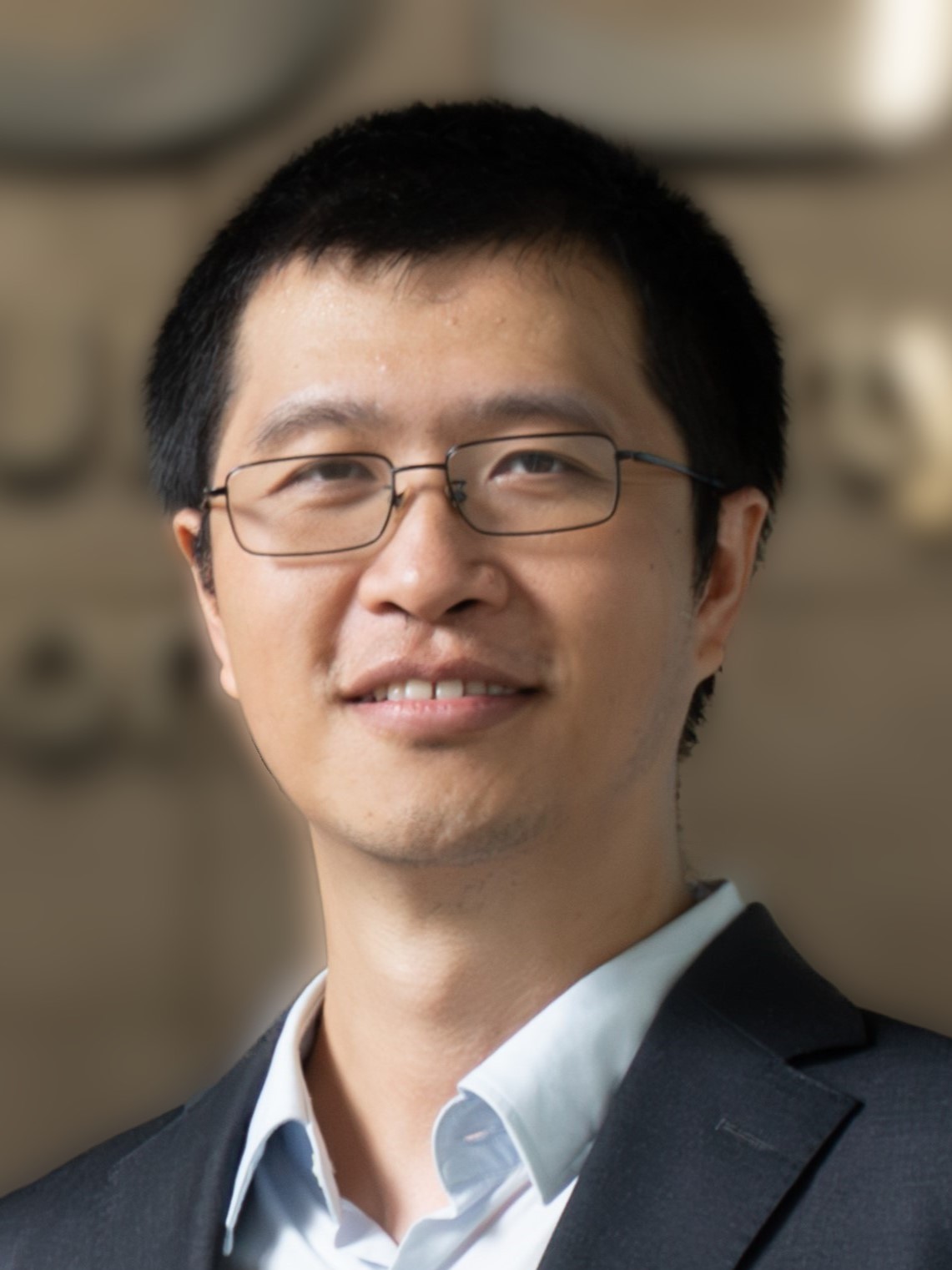}}]{Kaidi Yang} (Member, IEEE) is an Assistant Professor in the Department of Civil and Environmental Engineering at the National University of Singapore. Prior to this, he was a postdoctoral researcher with the Autonomous Systems Lab at Stanford University. He obtained a PhD degree from ETH Zurich and M.Sc. and B.Eng. degrees from Tsinghua University. His main research interest is the operation of future mobility systems enabled by connected and automated vehicles (CAVs) and shared mobility.
\end{IEEEbiography}

\vfill

\end{document}